\documentclass[preprint,12pt]{elsarticle}

\usepackage{amssymb}

\usepackage{url}
\usepackage{hyperref}
\usepackage{soul}
\usepackage{listings}
\usepackage{makecell}
\usepackage[dvipsnames]{xcolor}
\definecolor{backcolour}{rgb}{0.95,0.95,0.92}
\definecolor{premierEmerald}{HTML}{00b564}
\definecolor{premierEmerald+}{HTML}{006337}
\definecolor{premierPurple}{HTML}{7030A0}
\definecolor{premierPurple-}{HTML}{bc71f5}
\usepackage{tablefootnote}
\usepackage{tikz}
\usetikzlibrary{shapes.geometric}
\usepackage{amsmath}
\usepackage{mathtools}
\usepackage{booktabs}
\usepackage{pifont}
\usepackage[most]{tcolorbox}
\tcbuselibrary{skins,raster}
\usepackage{caption}
\usepackage{float}
\usepackage{etoolbox}
\usepackage{cancel}
\usepackage{enumitem} 
\usepackage{color}
\usepackage{threeparttable}

\journal{ArXiv}

\begin{document}

\begin{frontmatter}



\title{VERSE: Visual Embedding Reduction and Space Exploration – Clustering-Guided Insights for Training Data Enhancement in Visually-Rich Document Understanding}
\author[1]{Ignacio de Rodrigo\corref{cor1}}
\ead{iderodrigo@comillas.edu}
\cortext[cor1]{Corresponding author}

\author[1]{Alvaro J. Lopez-Lopez}
\ead{allopez@comillas.edu}

\author[1]{Jaime Boal}
\ead{jboal@comillas.edu}

\affiliation[1]{organization={Institute for Research in Technology, ICAI School of Engineering, Comillas Pontifical University},
            addressline={Calle Rey Francisco, 4}, 
            city={Madrid},
            postcode={28008}, 
            state={Madrid},
            country={Spain}}




\begin{abstract}
This work introduces VERSE, a methodology for analyzing and improving Vision–Language Models applied to Visually-rich Document Understanding by exploring their visual embedding space. VERSE enables the visualization of latent representations, supporting the assessment of model feasibility. It also facilitates the identification of problematic regions and guides the generation of synthetic data to enhance performance in those clusters. We validate the methodology by training on the synthetic MERIT Dataset and evaluating on its real-world counterpart, MERIT Secret. Results show that VERSE helps uncover the visual features associated with error-prone clusters, and that retraining with samples containing these features substantially boosts F1 performance without degrading generalization. Furthermore, we demonstrate that on-premise models such as Donut and Idefics2, when optimized with VERSE, match or even surpass the performance of SaaS solutions like GPT-4 and Pixtral.
\end{abstract}

\end{frontmatter}



\section{Introduction}
\label{sec:introduction}

Vision–Language Models (VLMs) have recently demonstrated remarkable capabilities in multimodal contexts, enabling them to perform complex tasks that combine visual and textual understanding. Within this family, models for \textit{Visually-rich Document Understanding} (VrDU) aim to interpret structured documents by integrating layout, textual, and visual features into a unified representation or \textit{embedding}. Understanding how these embeddings are organized, how they relate to one another, and what latent patterns emerge brings us closer to uncovering each model’s underlying \textit{semantic structure} \cite{spies2025transformers}. Moreover, optimizing the spatial arrangement of embeddings relative to their neighbors not only enhances pattern detection and interpretability in the latent space but also leads to improved performance in downstream tasks \cite{li2025multi}.

\begin{figure}
\centering
\includegraphics[width=1\columnwidth]{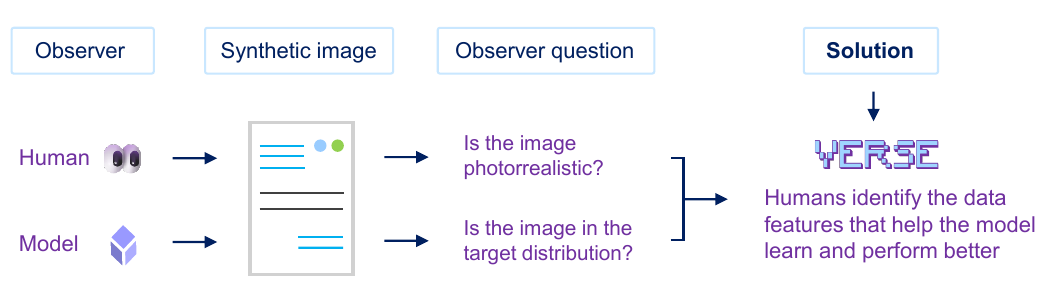}
\caption{Paradigm shift proposed in this work. Traditionally, the quality of synthetic images in a dataset is assessed from an anthropocentric perspective, answering the question of whether such images appear photorealistic. In contrast, this work proposes evaluating the images from the model’s perspective, which entails analyzing their visual embeddings to determine whether a synthetic image lies within the target distribution, as perceived by the model itself.}
\label{fig:introduction}
\end{figure}

VLMs are generally trained on large-scale, real-world multimodal corpora, but rely on synthetic data in domains where access to real samples is limited \cite{nassar2025smoldocling}. In such cases, visual quality is often evaluated from a human perspective, emphasizing whether the samples appear photorealistic or plausible. However, visual realism does not necessarily imply usefulness for the model itself. From the model’s perspective (its semantic structure representation), what truly matters is whether a synthetic sample lies within the same distribution as real samples in the visual–semantic embedding space. This perspective motivates a paradigm shift in the way synthetic training data are evaluated: rather than judging visual quality from an anthropocentric point of view, data quality should be assessed through the model’s own internal representations (see Figure~\ref{fig:introduction}).

To translate this idea into a practical methodology, we propose \textit{VERSE} \textit{(Visual Embedding Reduction and Space Exploration)}. VERSE is a methodology designed to analyze, interpret, and leverage the structure of the visual embedding space of VLMs. VERSE serves three main purposes: (i) \textit{model interpretability}, by assessing whether the visual embeddings of a given model form coherent patterns aligned with the target task; (ii) \textit{explainability and model boosting}, by identifying error-inducing regions in the embedding space and enriching training datasets with representative samples from these regions; and (iii) \textit{human-centered explainability}, aimed at revealing the visual and structural features that drive the patterns underlying model behavior and task performance.

The methodology is validated on the MERIT Dataset~\cite{de2025merit} and its real-world counterpart MERIT Secret, focusing on key information extraction as a sequence-generation task in Spanish.

Our main contributions are as follows:

\begin{itemize}
    \item We present a novel methodology for visual embedding analysis and latent-space visualization, named VERSE \textit{(Visual Embedding Reduction and Space Exploration)}.
    \item We employ VERSE to assess a model’s validation for a specific task (VrDU in this case) by examining the structure and patterns present in its \textit{reduced embedding space} (RES).
    \item For suitable models, VERSE identifies problematic clusters and helps to reveal which visual features should be modified to enhance performance within them.
    \item Applying VERSE yields a measurable performance boost, enabling on-premise models to match (and in some cases surpass) the results of leading SaaS-based solutions such as GPT-4 and Pixtral~\cite{agrawal2024pixtral}.
\end{itemize}

The remainder of this paper is organized as follows. Section~\ref{sec:related_work} reviews the main advances in VrDU, covering both model and dataset perspectives. Section~\ref{sec:methodology} introduces the VERSE methodology. Section~\ref{sec:results} presents the experimental evaluation, showing how VERSE improves performance by addressing error-inducing regions in the latent space and benchmarking on-premise models against SaaS-based solutions. Finally, Section~\ref{sec:discusion} discusses the results, and Section~\ref{sec:conclusions_contributions} concludes the paper by summarizing the contributions and outlining future directions.

\section{Related Work}
\label{sec:related_work}

The document analysis domain has emerged as one of the most impactful areas of applied Artificial Intelligence, attracting both industrial and academic attention. This growing interest has led to a broad range of research efforts (from model architectures to dataset generation) aimed at enhancing the performance and robustness of VrDU systems.

Among the most influential model families in this field is the LayoutLM \cite{xu2020layoutlm} series, which represents OCR-dependent VLMs. The original LayoutLM model was pre-trained on tasks such as document classification and form understanding, establishing the foundations for multimodal document analysis. LayoutLMv2 \cite{xu2021layoutlmv2} introduced additional pre-training objectives (text-image alignment and text-image matching) designed to improve the interaction between visual and textual modalities. Building upon this, LayoutXLM \cite{Xu2021} extended the approach to multilingual settings using a corpus covering 53 languages, and released the XFUND dataset \cite{Xu2022} as a cross-lingual benchmark. Later, LayoutLMv3 \cite{huang2022layoutlmv3} refined cross-modal representations by introducing a Word–Patch Alignment task, promoting stronger associations between visual patches and their corresponding textual tokens.

Despite the strong performance of this family, OCR dependence remains a critical limitation. On one hand, OCRs struggle with challenging real-world conditions such as complex layouts or degraded text \cite{palacios2008system}; on the other, they impose a rigid reading order that constrains token sequence modeling. XYLayoutLM \cite{gu2022xylayoutlm} explicitly addresses this issue by proposing a token reordering mechanism based on spatial coordinates. In contrast, recent approaches have moved toward eliminating the OCR dependency altogether. Donut \cite{kim2022ocr} introduced an end-to-end OCR-free architecture, marking a paradigm shift in document understanding. Building upon this direction, DocParser \cite{dhouib2023docparser} enhances character-level discrimination, while UDOP \cite{tang2023unifying} broadens the modeling scope by incorporating inpainting capabilities for coherent text editing and region reconstruction in document layouts.

More recently, an alternative line of work has focused on instruction-tuned multimodal models. These architectures extend LLMs with visual encoders and cross-modal fusion mechanisms, enabling end-to-end reasoning across text and images. PaliGemma \cite{beyer2024paligemma} offers an efficient, open-source variant of PaLI \cite{chenpali}, combining a SigLIP \cite{zhai2023sigmoid} encoder with a Gemma-based \cite{team2024gemma} decoder. LLaVA \cite{liu2023visual} integrates CLIP \cite{radford2021learning} with a Vicuna \cite{vicuna2023} backbone, emphasizing conversational and general-purpose multimodal tasks. In contrast, Idefics2 \cite{laurenccon2024matters} and Idefics3 \cite{laurenccon2024building} provide larger-scale, instruction-aligned systems trained on diverse image–text corpora.

Finally, all these developments in VrDU tasks have led to the emergence of toolkits such as Docling \cite{livathinos2025docling}, developed by IBM, which illustrates the growing industrial interest in this field. This ecosystem is also accelerating the development of new ad hoc strategies for document processing (such as table tokenization \cite{lysak2023optimized}) and fostering the release of open-source models like Smol-Docling \cite{nassar2025smoldocling}. These advances exemplify the increasing collaboration between industry and the open-source community, highlighting the pivotal contribution of Hugging Face to the deep learning ecosystem.

In parallel with the rapid development of VLMs, numerous datasets have been introduced to support progress in VrDU for multiple tasks. The NNE dataset \cite{ringland2019nne} focuses on Named Entity Recognition (NER), providing finely nested labels that enrich the semantic and syntactic context of each token. FUNSD \cite{Jaume2019} is another widely used dataset for token classification, consisting of real scanned documents annotated with text, layout, and semantic information in English. To overcome the linguistic limitation of previous corpora, XFUND \cite{Xu2022} extends this approach to include samples in up to seven languages, offering a multilingual benchmark for VrDU.

Domain-specific datasets have also emerged. CORD \cite{Park2019} and SROIE \cite{huang2019icdar2019} target purchase receipt analysis, supporting tasks such as text localization and key information extraction. PublayNet \cite{zhong2019publaynet} focuses on document layout analysis, providing digitally born images of scientific articles. However, since these are not scanned or photographed, a domain gap arises when applying trained models to real-world documents. Similarly, DocVQA \cite{mathew2021docvqa} and its extensions (PDF-VQA \cite{ding2023vqa}, SlideVQA \cite{tanaka2023slidevqa}, and InfographicVQA \cite{mathew2022infographicvqa}) expand document understanding into visual question answering, multi-page reasoning, and numerical inference tasks, pushing VrDU toward more complex multimodal reasoning challenges.

Other approaches have aimed to minimize human intervention in the dataset creation process. For instance, CRAFT \cite{ziegler2024craft} enables the construction of textual datasets in a few-shot setting, where representative samples are used to retrieve semantically related examples from large corpora via embedding-based similarity. In cases where no suitable examples exist, controlled data generation methodologies (such as MERIT \cite{de2025merit}) are essential for building multimodal datasets with highly structured content.

\section{Methodology}
\label{sec:methodology}

\begin{figure}
\centering
\includegraphics[width=1\columnwidth]{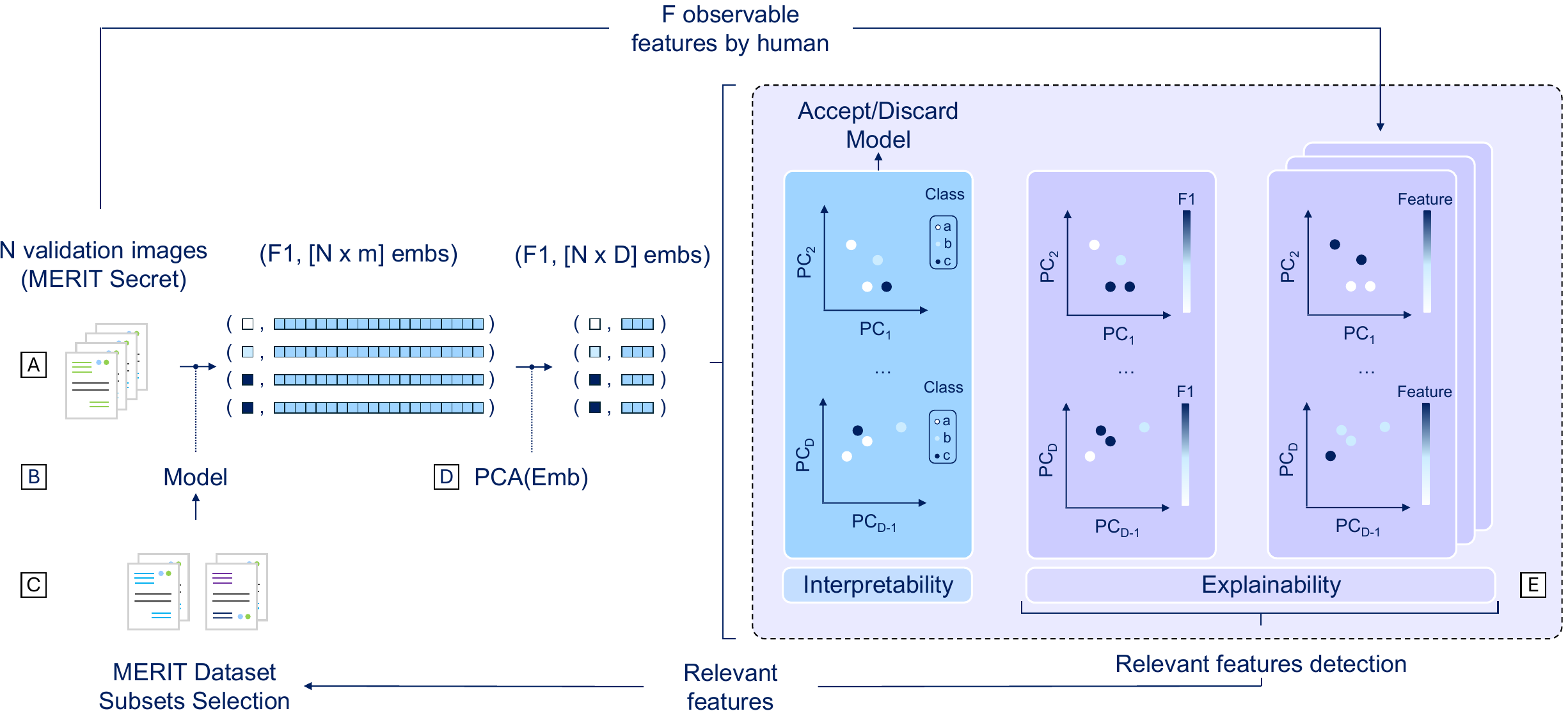}
\caption{VERSE methodology components. The validation dataset \textit{MERIT secret} (A) is processed by models' visual encoders (B) to obtain \textit{visual embeddings}. F1 scores are obtained after inference on the validation dataset with fine-tuned models on the different \textit{MERIT Dataset} versions. High-dimensional visual embeddings are reduced to a lower-dimensional space (D), known as the \textit{Reduced Embedding Space}. This space provides better model \textit{interpretability}, while overlaying the samples' visual features and the F1 scores enhances model \textit{explainability} (E). Sections \ref{subsec:verse_blocks}.\ref{subsec:test-dev_dataset} to \ref{subsec:verse_blocks}.\ref{subsec:clustering} explain in further detail the components (A-E) involved in the methodology.}
\label{fig:methodology}
\end{figure}

The methodology proposed in this paper, VERSE, builds upon the inspection, reduction, and visual analysis of visual embeddings in VLMs with three main objectives: (i) \textit{interpretability} to assess a model’s potential for executing a given task (sequence generation in this case) in Subsection \ref{subsec:methodology_model_suitability}; (ii) from a human-centered perspective, \textit{explainability} to uncover the relevant features embedded in the synthetic training samples in Subsection \ref{subsec:methodology_explainability}; and (iii) \textit{explainability} to identify model-specific error-inducing features and use them to construct enriched training datasets that enhance model performance in Subsection \ref{subsec:methodology_model_boosting}. The overall workflow of the VERSE methodology is depicted in Figure~\ref{fig:methodology} and its building blocks are explained in Subsections \ref{subsec:verse_blocks}.\ref{subsec:test-dev_dataset}-\ref{subsec:verse_blocks}.\ref{subsec:clustering}.

\subsection{VERSE building blocks}
\label{subsec:verse_blocks}

\begingroup
\renewcommand{\thesubsubsection}{\Alph{subsubsection}}

\subsubsection{Validation dataset}
\label{subsec:test-dev_dataset}

The validation dataset corresponds to MERIT Secret, a dataset of real-world samples provided to the authors under NDA. It comprises 152 images of student transcripts of records in Spanish, captured with different devices and under varying lighting conditions. The samples originate from 10 different schools, each with its own template and formatting conventions, though minor variations may also appear within the same institution.

The dataset characteristics can be grouped into two categories: (i) intrinsic document properties, such as the arrangement of elements (tables, headings, images), or the representation of grades (numeric or alphanumeric); and (ii) environmental factors, external to the document itself, including wrinkles, shadows, or artifacts introduced during the anonymization process. As Figure \ref{fig:methodology} shows, MERIT Secret samples are labelled attending to these human-observed features. Table \ref{tab:metadata-condensed} exposes and describes them.

Figure~\ref{fig:merit_secret} illustrates the distribution of samples across schools and provides, in a privacy-preserving manner, a thumbnail for each subset that conveys the structural and visual 	characteristics of the documents.

The validation dataset serves as the source for several data signals in the VERSE methodology: (i) the F1 score computed for the 152 samples; (ii) the corresponding embedding produced by models for every image; and (iii) the observable features extracted as human-interpretable labels.

\begin{figure}
\centering
\includegraphics[width=1\columnwidth]{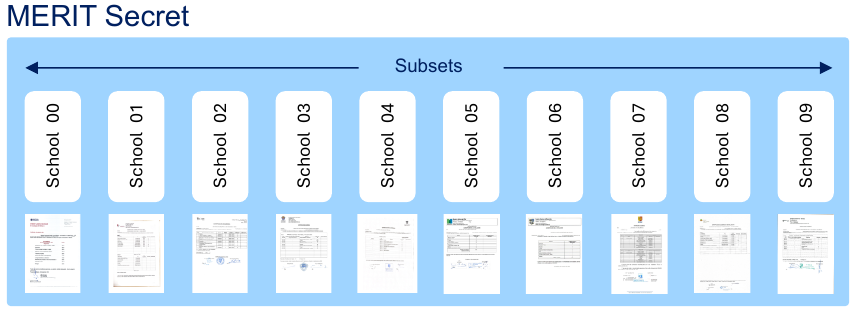}
\caption{Validation dataset: MERIT Secret, a dataset with real and anonymized samples, with 10 categories subdivided by school. Table \ref{tab:metadata-condensed} shows the relevant features extracted in MERIT Secret.}
\label{fig:merit_secret}
\end{figure}

\subsubsection{Models}
\label{subsec:models}

The models used are VLMs available on Hugging Face. These models are based on the transformer architecture and incorporate a visual encoder to process both text and images as input. Table~\ref{tab:vlm_visual_encoders} summarizes their main features.

The visual embeddings for each model are extracted from the \textit{last hidden state} of the visual block. All training runs performed with the different versions of the training dataset are conducted with the visual backbone frozen, ensuring that embedding comparisons and dimensionality reduction are carried out over a consistent visual semantic structure. Otherwise, each fine-tuning process would introduce changes in the underlying visual representation.

Within the VERSE methodology, the models are involved in two key stages: (i) obtaining the visual embedding for each sample in the validation set from the off-the-shelf model, and (ii) computing the F1 score for each sample using the fine-tuned model.


\begin{table}[ht]
\centering
\tiny
\begin{threeparttable}
\begin{tabular}{l l l l}
\toprule
\makecell[tl]{\textbf{Model/} \\\textbf{HF repo ID}} &
\makecell[tl]{\textbf{Visual Encoder}\\\textbf{(VE)}} &
\makecell[tl]{\textbf{Is VE frozen}\\\textbf{in model P-T?}} &
\makecell[tl]{\textbf{Model P-T}\\\textbf{Datasets}}\\  
\midrule
\makecell[tl]{\textbf{Donut} \cite{kim2022ocr} \\ naver-clova-ix/donut-base} & \makecell[tl]{Swin Transformer \cite{liu2021swin}} & False & Synth-Dog \cite{kim2022ocr} docs\\
\makecell[tl]{\textbf{Idefics2} \cite{laurenccon2024matters} \\ HuggingFaceM4/idefics2-8b} & \makecell[tl]{SigLIP\,ViT-So400M \cite{zhai2023sigmoid}} & \makecell[tl]{False,\\ LoRA adaption} & \makecell[tl]{OBELICS \cite{laurenccon2023obelics}\tnote{a}\\ OCR-IDL \cite{biten2022ocr}, PDFA.}\\
\makecell[tl]{\textbf{PaliGemma} \cite{beyer2024paligemma} \\ google/paligemma-3b-pt-224} & \makecell[tl]{SigLIP\,ViT-So400M \cite{zhai2023sigmoid}} & False & \makecell[tl]{WebLI \cite{beyer2024paligemma}, CC3M-35L \cite{sharma2018conceptual},\\ OpenImages \cite{kuznetsova2020open}, WIT \cite{srinivasan2021wit},\\ VQ2A-CC3M \cite{changpinyo2022all}\tnote{b}} \\
\makecell[tl]{\textbf{LLaVA} \cite{liu2023visual} \\ LLaVA-hf/LLaVA-1.5-7b-hf} & \makecell[tl]{CLIP\,ViT-L/14 \cite{radford2021learning}} & True & N/A \\
\bottomrule
\end{tabular}
\begin{tablenotes}\tiny
\item[a] Interleaved image–text documents from OBELICS.
\item[b] Authors use specific subsets, but they do not specify what their content is.
\end{tablenotes}
\end{threeparttable}
\caption{Summary of Visual Encoders (VE). Models whose visual encoders are unfrozen (Donut and Idefics2) and pre-trained (P-T) on document-like datasets tend to produce richer and more structured visual embedding spaces, as shown in Figure~\ref{fig:reduced_space}. In contrast, PaliGemma relies more heavily on general-purpose, object-centric pre-training data, which is less aligned with document-specific visual cues.}
\label{tab:vlm_visual_encoders}
\end{table}

\subsubsection{Training dataset}
\label{subsec:training_dataset}
The training dataset corresponds to the Spanish partition of the MERIT Dataset. To analyze how the models respond to different visual stimuli, we generate different versions from MERIT vanilla (Figures~\ref{fig:training_dataset}.C and D). These data augmented versions introduce additional granularity along the vanilla spectrum of samples. Figure~\ref{fig:training_dataset}.A illustrates this expansion of dataset versions and their relative positioning (from a qualitative perspective) according to the increasing amount and quality of visual information. Table~\ref{tab:training_dataset_versions} summarizes the characteristics of each version. Figure~\ref{fig:training_dataset}.B depicts, for a given version, the subdivision of the dataset into subsets organized by school. Each subset, corresponding to a specific institution, exhibits distinctive visual, layout, and textual features. These characteristics show varying degrees of alignment with those of the validation dataset (MERIT Secret).

In this context, within the VERSE methodology, each model is iteratively fine-tuned using a specific version or composition of the MERIT dataset as training data.

\begin{figure}
\centering
\includegraphics[width=1\columnwidth]{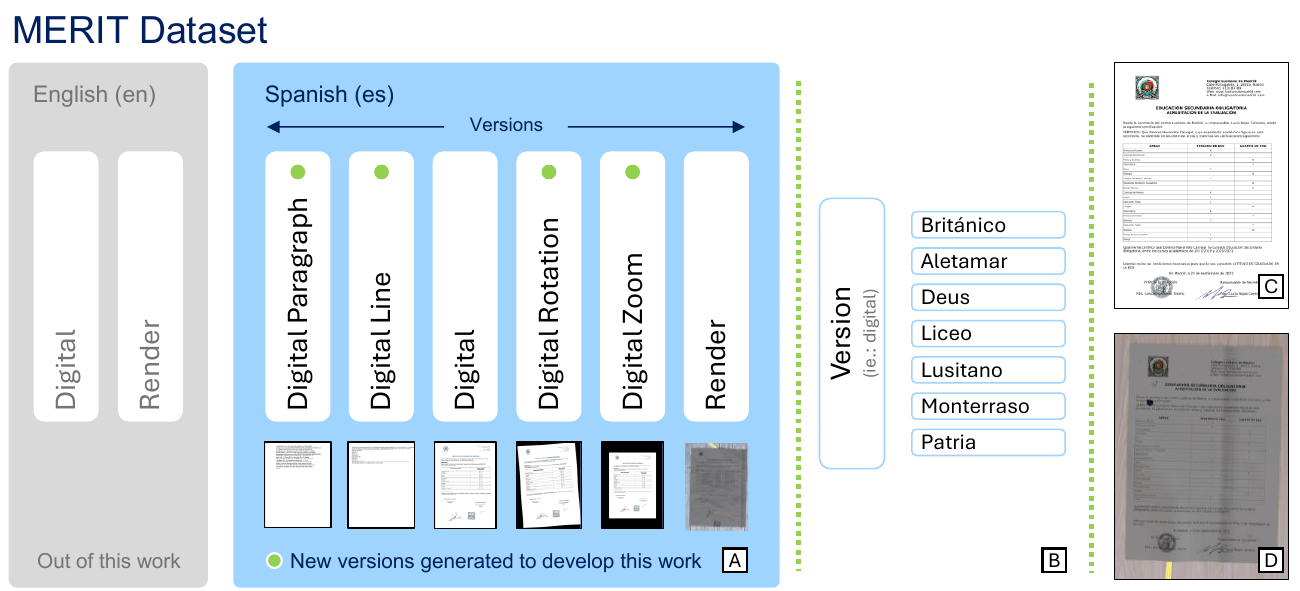}
\caption{Training samples used. We employ the Spanish-language subsets of the MERIT Dataset, across its different versions (A). These versions are detailed in Table \ref{tab:training_dataset_versions}). Each version comprises data from seven different schools (B). New versions complement the vanilla MERIT Dataset, composed of digital document samples (C) and their renderized versions (D).}
\label{fig:training_dataset}
\end{figure}

\begin{table}[ht]
\centering
\scriptsize 
\begin{tabular}{l p{5cm} l}
\toprule
\textbf{Version} & \textbf{Description}  & \textbf{MERIT Dataset key} \\ 
\midrule
Digital Paragraph & Plain text, no structure or visual features (single paragraph). & es-digital-paragraph-degradation-seq \\
Digital Line & Line breaks per \textit{subject–grade} pair; minimal structure, no visuals. & es-digital-line-degradation-seq \\
Digital & Base version; includes structure and visuals (headings, stamps, signatures). & es-digital-seq \\
Digital Rotation & Rotated samples; black padding fills corners. & es-digital-rotation-degradation-seq \\
Digital Zoom & Zoomed-out samples; black padding surrounds content. & es-digital-zoom-degradation-seq \\
Rendered & Blender-rendered samples with 3D mesh, lighting, background, textures. & es-render-seq \\
\bottomrule
\end{tabular}
\caption{Overview of the visual features incorporated in the different versions of the MERIT dataset.}
\label{tab:training_dataset_versions}
\end{table}

\subsubsection{Embedding reduction}
\label{subsec:embedding_reduction}

In VERSE, embeddings are extracted from the off-the-shelf (pre-trained) versions of the models. The visual encoders in VLMs process each input image by dividing it into [n$\times$m] patches, which are then progressively reduced to obtain their latent representations. This reduction may be accompanied by a re-projection into the input space of the textual encoder to meet its dimensionality requirements. Conceptually, each patch is processed, resulting in a [n$\times$m$\times$L] tensor. In VERSE, an average pooling operation is applied to reduce this tensor to a final vector of size [1$\times$L], so that each image is condensed into a single embedding (thus obtaining a model-specific representation for each sample). Figure \ref{fig:embedding_obtention} shows the embedding obtention process. We use the frozen off-the-shelf version of the models to produce these embeddings.

\begin{figure}
\centering
\includegraphics[width=0.9\columnwidth]{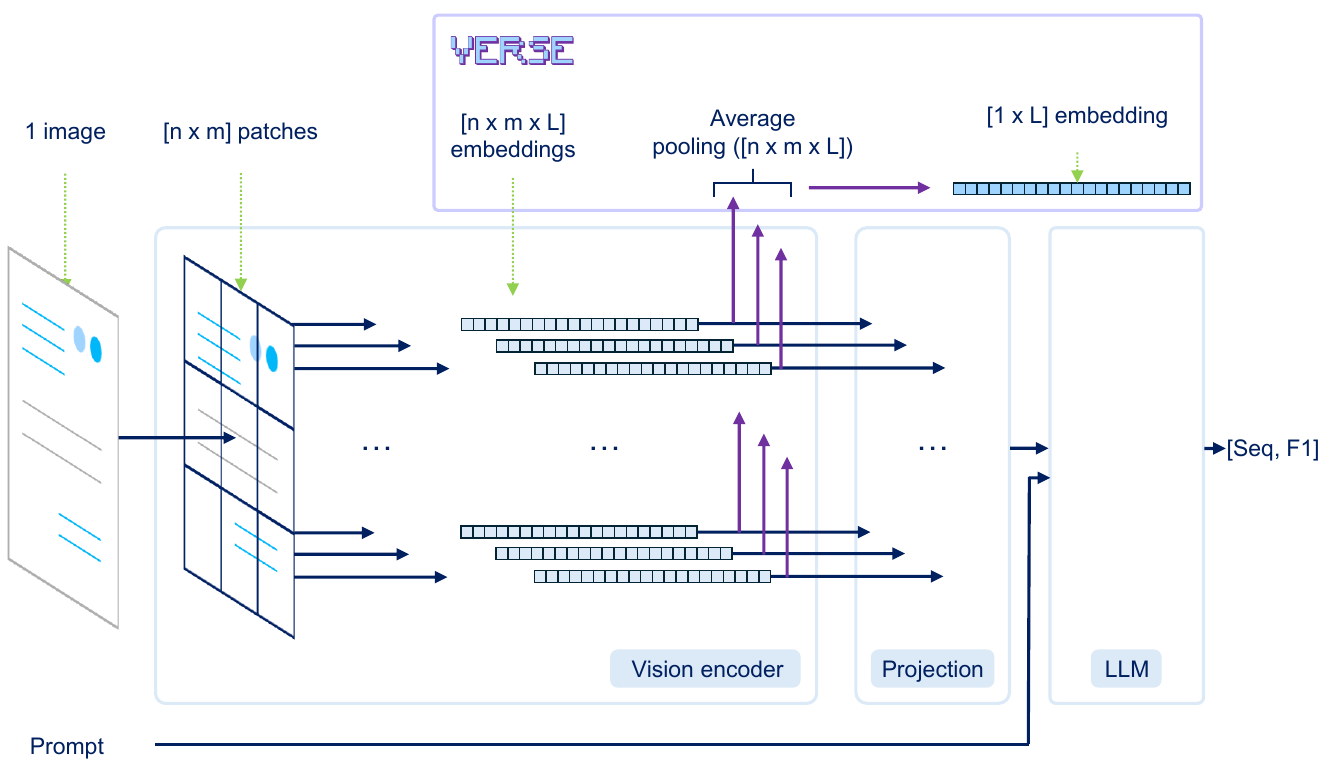}
\caption{Qualitative illustration of the pipeline used to obtain a single visual embedding per image. The procedure is encoder-specific: Donut, PaliGemma, and LLaVA employ a lightweight MLP projection to align visual and textual embedding dimensions, whereas Idefics2 incorporates a more advanced projection mechanism within the vision encoder. For consistency across models, we extract visual embeddings from the output of the vision block in all cases.}
\label{fig:embedding_obtention}
\end{figure}

The visual embeddings used in this work have L-dimensional representations on the order of several thousand. To obtain a human-interpretable graphical representation, these embeddings must therefore be reduced. Figure~\ref{fig:reduced_space} shows the Reduced Embedding Spaces (RES) for the Idefics2 \cite{laurenccon2024matters} (A) and LLaVA \cite{liu2023visual} (B) models. Table~\ref{tab:embeddings_summary} summarizes the main characteristics of the embeddings produced by each model.

\begin{figure}
\centering
\includegraphics[width=1\columnwidth]{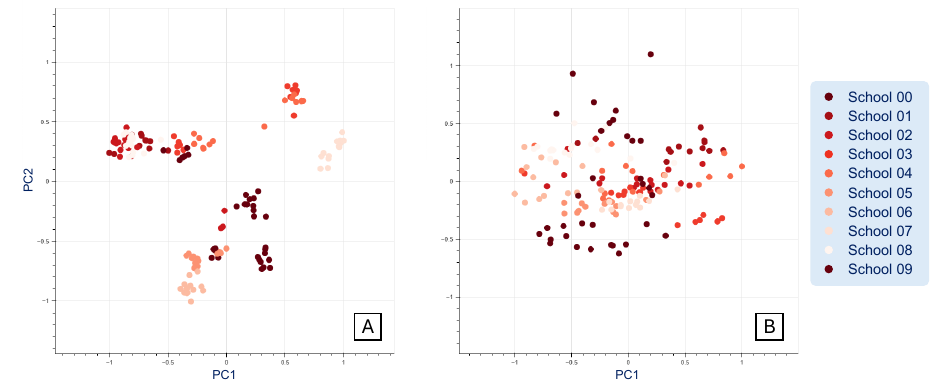}
\caption{Graphical representation (PC1 vs. PC2) of the Reduced Embedding Space (RES) obtained for Idefics2 \cite{laurenccon2024matters} (A) and LLaVA \cite{liu2023visual} (B). The spaces were computed by applying PCA to the 152 embeddings from the validation dataset. VERSE enables a visual assessment of the model’s visual-world representation quality: well-defined and separable clusters (A) indicate stronger representational capacity and higher task potential, whereas a disorganized or highly entangled RES (B) reflects weaker visual discrimination and poorer downstream performance.}
\label{fig:reduced_space}
\end{figure}

\begin{table}[h]
\centering
\tiny 
\begin{tabular}{l l l l l}
\toprule
\makecell[tl]{\textbf{Model}} & 
\makecell[tl]{\textbf{Visual}\\\textbf{Encoder}} & 
\makecell[tl]{\textbf{Visual Output}\\\textbf{Dimensionality [L]}} &
\makecell[tl]{\textbf{Visual-to-Text}\\ \textbf{Projection}} &
\makecell[tl]{\textbf{Projection Module}\\\textbf{Location}} \\
\midrule
Donut \cite{kim2022ocr} & Swin Transformer \cite{liu2021swin} & 1024 & Linear & Outside visual module\\
Idefics2 \cite{laurenccon2024matters} & SigLIP ViT-So400M \cite{zhai2023sigmoid} & $1152 \Rightarrow 4096$ & MLP + Perceiver & Within visual module\\
PaliGemma \cite{beyer2024paligemma} & SigLIP ViT-So400M \cite{zhai2023sigmoid} & 1152 & Linear & Outside visual module\\
LLaVA \cite{liu2023visual} & CLIP ViT-L/14 \cite{radford2021learning} & 1024 & Linear & Outside visual module\\
\bottomrule
\end{tabular}
\caption{Summary of visual encoders (VE) and their embedding outputs across the evaluated models. Idefics2 \cite{laurenccon2024matters}, which incorporates its projection mechanism directly within the \textit{visual module}, produces richer and more structured visual embeddings (see Figure~\ref{fig:reduced_space}). All embeddings were extracted as the \textit{last hidden state} of each model's visual module.}
\label{tab:embeddings_summary}
\end{table}

\subsubsection{Clustering}
\label{subsec:clustering}

As a result of dimensionality reduction, models with sufficiently rich visual representations for the target task (sequence generation for Visually-rich Document Understanding, VrDU) exhibit well-defined clustered regions within the embedding space (Figure \ref{fig:reduced_space}). Analyzing these clusters deepens model \textit{interpretability} (Section~\ref{subsec:methodology_model_suitability}), while overlaying complementary information (such as F1 scores or the visual and structural attributes summarized in Table~\ref{tab:metadata-condensed}) enhances \textit{explainability} (Sections~\ref{subsec:methodology_model_boosting} and~\ref{subsec:methodology_explainability}).

\endgroup

\subsection{Interpretability and model feasibility}
\label{subsec:methodology_model_suitability}

Dimensionality reduction through PCA reveals distinct clustering patterns. The degree of cluster definition varies depending on the model from which the embeddings are derived. Figure~\ref{fig:reduced_space} illustrates the clusters projected onto the first two principal components for Idefics2 \cite{laurenccon2024matters} (A) and LLaVA \cite{liu2023visual} (B). Since we are working within a highly specific domain, we are not interested in general or isotropic embedding spaces~\cite{balestriero2025lejepa}, but rather in a highly structured space that leverages dominant directions.

To validate that the structure of the embedding space is preserved after dimensionality reduction, trustworthiness and proximity metrics were computed between the original high-dimensional embeddings and their PCA-reduced projections. Table~\ref{tab:model_suitability} reports both metrics, together with the Silhouette score (computed over the reduced embedding space) for each model.

\begin{table}[ht]
\centering
\tiny
\begin{tabular}{lcccclcc}
\toprule
\makecell[tl]{\textbf{Model}\\\textbf{}} & 
\makecell[tl]{\textbf{Trust.}\\\textbf{$\in [0, 1]$}} &
\makecell[tl]{\textbf{Prox.}\\\textbf{$\in [0, 1]$}} &
\makecell[tl]{\textbf{K-Means}\\\textbf{clusters}} &
\makecell[tl]{\textbf{Radius}\\\textbf{$\mu, [min, max]$}} &
\makecell[tl]{\textbf{Density}\\\textbf{$\mu, [min, max]$}} &
\makecell[tl]{\textbf{Silh.}\\\textbf{$\in [-1, 1]$}} &
\makecell[tl]{\textbf{RES}\\\textbf{}}\\  
\midrule
Idefics2 \cite{laurenccon2024matters} & 0.96 & 0.98 & 7& 0.30 [0.21–0.47]& 243 [69–476]&  0.63 & $\checkmark$\\
Donut \cite{kim2022ocr} & 0.99 & 0.99 & 4& 0.46 [0.36–0.58]& 98 [76–150]& 0.50 & $\checkmark$\\
PaliGemma \cite{beyer2024paligemma} & 0.93 & 0.95 & 5& 0.40 [0.32–0.57]& 159 [19–253]&  0.38 & $\times$\\
LLaVA \cite{liu2023visual} & 0.97 & 0.97 & 5& 0.58 [0.42–1.02]& 51 [8–72]& 0.35 &  $\times$\\
\bottomrule
\end{tabular}
\caption{Quantitative analysis of the Reduced Embedding Space (RES).
\textit{Trust.} (Trustworthiness) and \textit{Prox.} (Proximity) measure how well the local neighborhood structure is preserved after dimensionality reduction. \textit{K-Means clusters} indicates the number of clusters automatically detected. \textit{Radius} represents the average and range of cluster radii, providing a measure of cluster compactness, while \textit{Density} reflects data concentration. \textit{Silh.} (Silhouette score) quantifies the degree of cluster separability: higher values indicate that samples are well matched to their own cluster and distant from others, meaning a more structured and discriminative embedding space. Finally, \textit{RES} summarizes the qualitative richness of the model’s visual embedding space. Based on these metrics (especially the Silhouette score) Idefics2 \cite{laurenccon2024matters} and Donut \cite{kim2022ocr} exhibit well-structured and semantically meaningful embedding spaces, whereas PaliGemma \cite{beyer2024paligemma} and LLaVA \cite{liu2023visual} display weaker cluster separation and lower representational quality.}
\label{tab:model_suitability}
\end{table}

VERSE proposes a qualitative and quantitative inspection of the reduced embedding space to assess a model’s potential to solve the target task. A better cluster definition implies that the models can distinguish samples based on the discriminative features, which is interpreted as a richer understanding of the problem. Idefics2 \cite{laurenccon2024matters} and Donut \cite{kim2022ocr} are identified as the most suitable models, while PaliGemma \cite{beyer2024paligemma} and LLaVA \cite{liu2023visual} are ruled out at this stage.

\subsection{Explainability and exploration of the reduced embedding space (RES)}
\label{subsec:methodology_explainability}

The projection performed by PCA entails an intrinsic loss of information and a high degree of entanglement among the principal components of the RES. Moving along a single component does not correspond to variations in one specific feature but rather in a combination of several. Nevertheless, overlaying visual features on the RES reveals clearly distinguishable clusters, with no apparent mixing of samples that exhibit different visual properties. This behavior indicates that the models (Donut \cite{kim2022ocr} and Idefics2 \cite{laurenccon2024matters}) are able to represent the visual structure of the data coherently, suggesting that their visual representations are sufficiently rich for the target task.

\begin{figure}
\centering
\includegraphics[width=1\columnwidth]{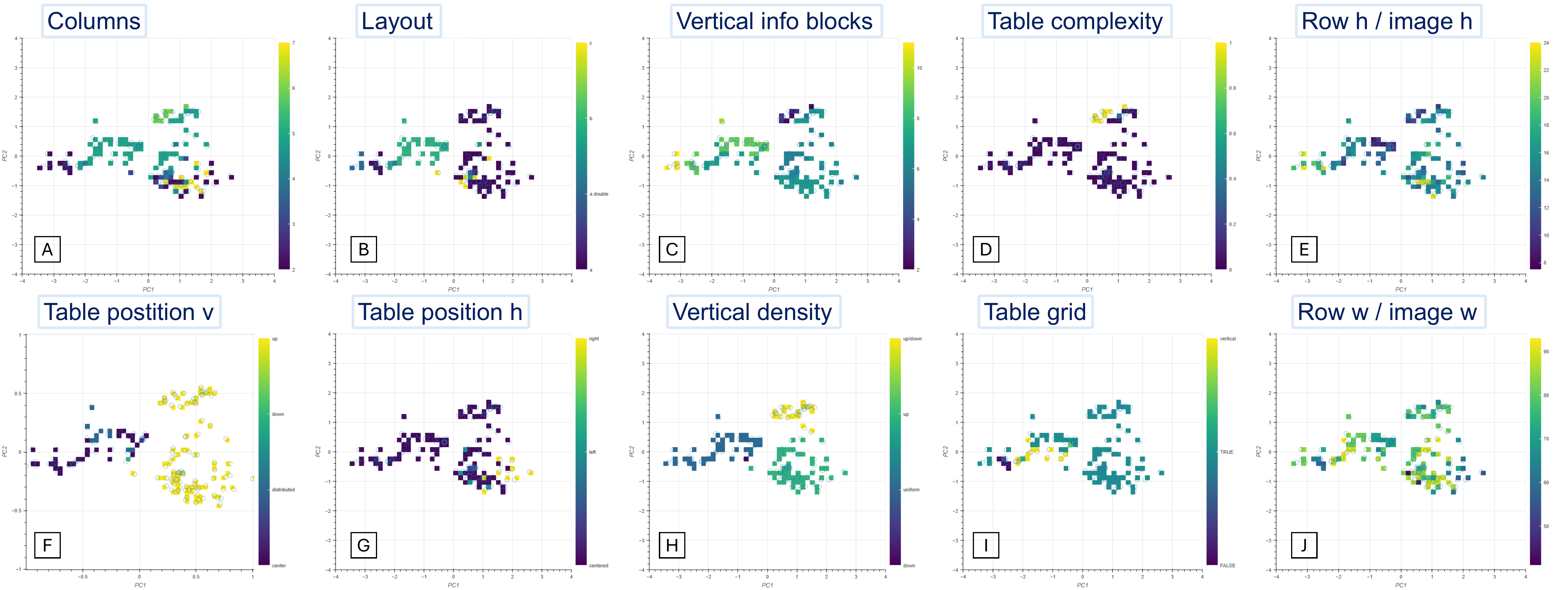}
\caption{Representation of visual features projected onto Donut’s \cite{kim2022ocr} RES. Each subplot shows a specific document feature distribution across PC1 vs. PC2. Number of \textit{columns} (A), \textit{layout} type (B), number of \textit{vertical information blocks} (C), \textit{table complexity} (D), \textit{row height / image height} ratio (zoom) (E), \textit{table vertical} (F) and \textit{horizontal} (G) \textit{position}, \textit{vertical density} (H), \textit{table grid} presence (I), and \textit{row width / image width} ratio (zoom)(J).}
\label{fig:metadata_pca_donut}
\end{figure}

Figure~\ref{fig:metadata_pca_donut} shows some visual features and their distribution across the Donut's \cite{kim2022ocr} RES. Most of these features are discrete and describe macro-visual properties of the documents, such as the number of columns (Figure~\ref{fig:metadata_pca_donut}.A), the vertical arrangement of information blocks (Figure~\ref{fig:metadata_pca_donut}.C), or the table position within the page (Figure~\ref{fig:metadata_pca_donut}.F and Figure~\ref{fig:metadata_pca_donut}.G). These features appear well-defined in the RES, indicating that the model can distinguish relevant structural patterns. On the other hand, for features not related to macroscopic document structure (such as zoom level,  Figures~\ref{fig:metadata_pca_donut}.E and J) show a higher degree of overlap between regions. Despite this, Donut \cite{kim2022ocr} is still capable of identifying low-zoom areas, which later correspond to low-F1 regions (see Section~\ref{subsec:methodology_model_boosting}).

\begin{figure}
\centering
\includegraphics[width=1\columnwidth]{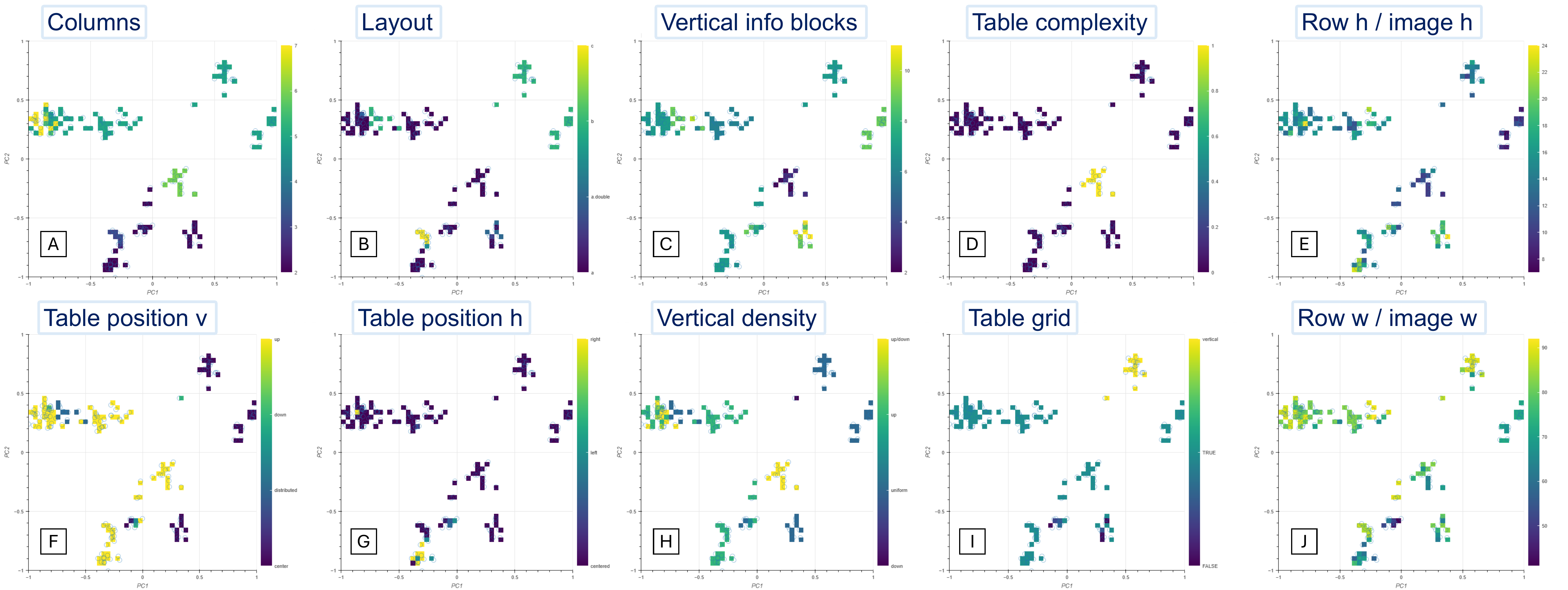}
\caption{Representation of visual features projected onto Idefics2’s \cite{laurenccon2024matters} RES. Each subplot shows a specific document feature distribution across PC1 vs. PC2. Features exposed are the same as the ones in Figure \ref{fig:metadata_pca_donut}.}
\label{fig:metadata_pca_idefics2}
\end{figure}

Figure~\ref{fig:metadata_pca_idefics2} illustrates the same analysis for Idefics2 \cite{laurenccon2024matters}, using the same set of features considered for Donut \cite{kim2022ocr}. Again, macro-level features (such as the number of columns or the vertical distribution of information, Figures ~\ref{fig:metadata_pca_idefics2}.A and C) enable the formation of well-defined clusters, while features related to zoom exhibit a more entangled behavior.

Overall, despite the high degree of feature entanglement present in principal components, both reduced spaces show that each cluster is locally driven by a specific combination of features. This behavior reinforces the interpretation that the model organizes its embedding space according to coherent visual patterns, mostly governed by macroscopic document layout properties. Consequently, the most influential features are those that describe the macrostructural design of the document (such as column arrangement, table position, or layout type) whereas lower-level details (e.g., the presence of stamps or signatures) have a lesser impact on the organization of the reduced spaces.

\subsection{Explainability and model boosting}
\label{subsec:methodology_model_boosting}

\begin{figure}
\centering
\includegraphics[width=1\columnwidth]{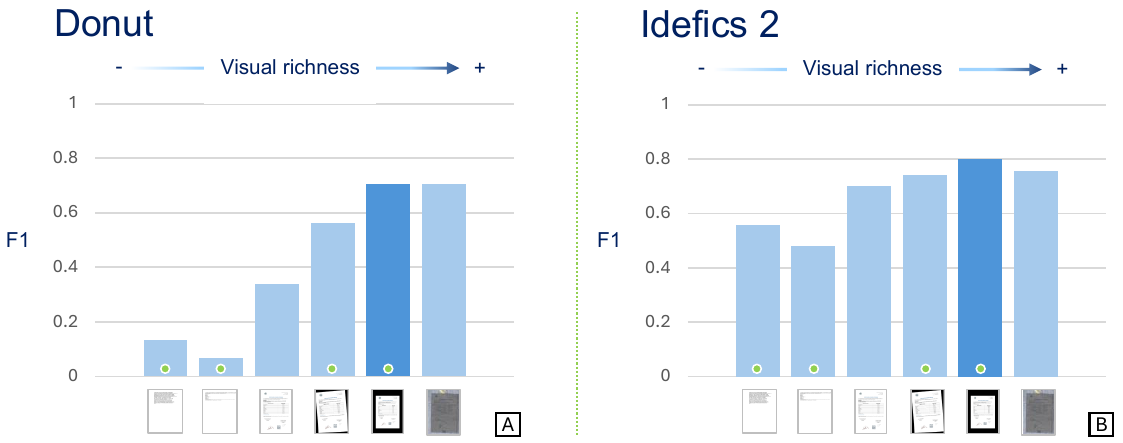}
\caption{Inference results for Donut \cite{kim2022ocr} (A) and Idefics2 \cite{laurenccon2024matters} (B). For each pre-trained model, the horizontal axis shows the different versions of the training dataset ordered by increasing visual richness (new versions include a green mark), while the vertical axis represents the F1 score obtained for each fine-tuned version. Both models exhibit clear sensitivity to the increasing level of visual richness.}
\label{fig:inference_results}
\end{figure}

VERSE first performs a training sweep in which the models are fine-tuned using progressively richer visual versions of the MERIT dataset (see Figure~\ref{fig:training_dataset}). Figure~\ref{fig:inference_results} shows the corresponding F1 scores on the validation dataset (MERIT Secret), for each fine-tuned version of Donut \cite{kim2022ocr} (A) and Idefics2 \cite{laurenccon2024matters} (B), ordered from the lowest to the highest level of visual richness (perceived by human).

\begin{figure}
\centering
\includegraphics[width=1\columnwidth]{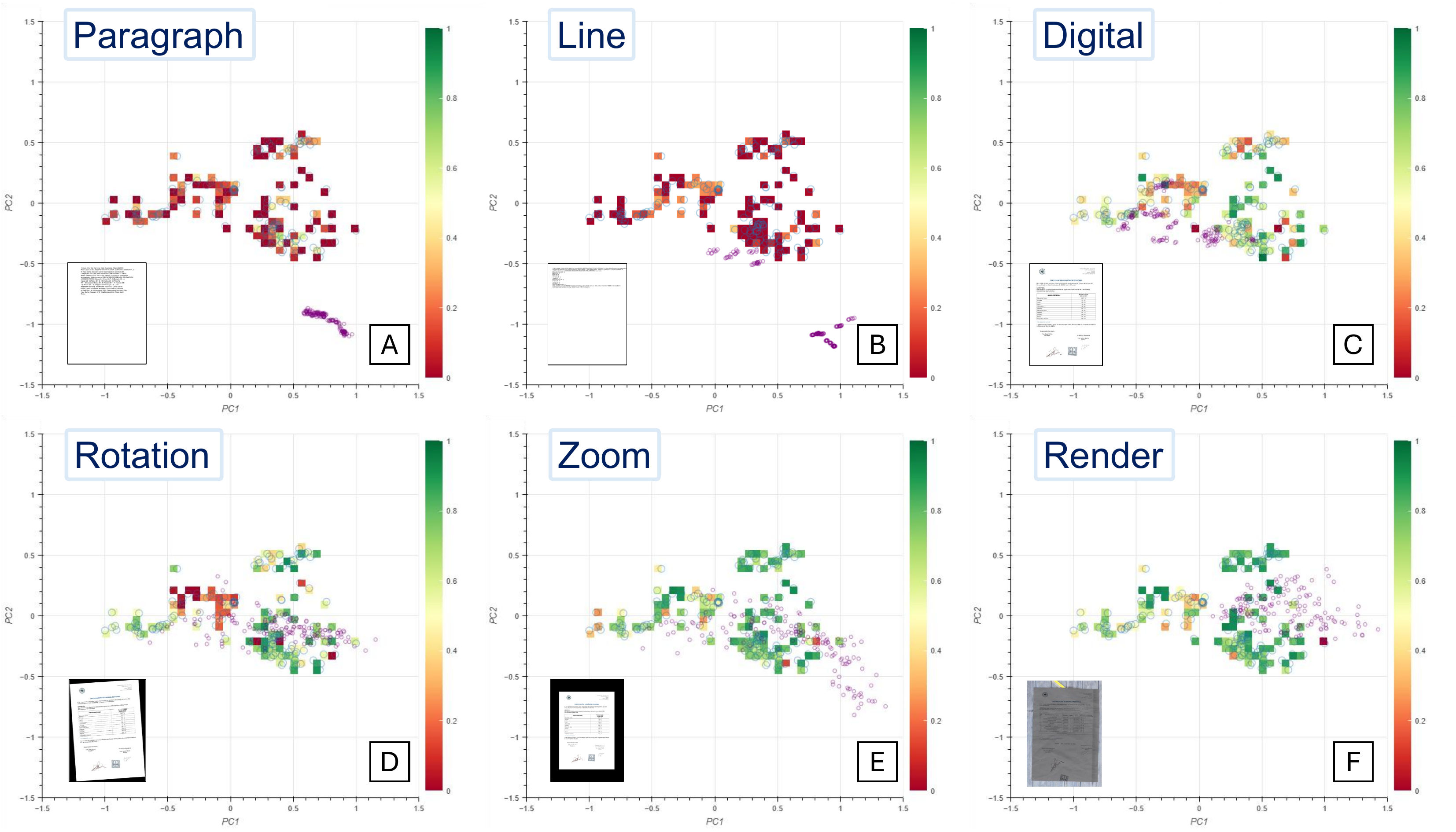}
\caption{Reduced Embedding Space (RES) for Donut \cite{kim2022ocr} (PC1 vs. PC2). Higher F1 values are shown in green, while lower ones appear in red. Each subplot includes, in the lower-left corner, a representative training sample used for fine-tuning and a subset of training examples shown in purple. The visual richness of the training data increases progressively from A to F, causing the corresponding embeddings to transition from concentrated regions to a more dispersed distribution across the RES.}
\label{fig:res_donut}
\end{figure}

Figure~\ref{fig:inference_results} shows an important first insight: models achieve competent performance when trained with samples that do not replicate human interpretation of photorrealism (\textit{rendered} version) but extract relevant information from versions including simple data augmentation techniques (\textit{rotation} and specially, \textit{zoom}). On the other hand, although Figure~\ref{fig:inference_results} illustrates the sensitivity of the different Donut \cite{kim2022ocr} and Idefics2 \cite{laurenccon2024matters} versions (trained with progressively richer visual inputs) when evaluated on the validation set, it does not reveal (i) how individual samples are distributed with respect to their F1 scores, or (ii) whether specific features are associated with poor performance. To address this, VERSE analyzes the visual embeddings and their representation within the reduced embedding space (RES). Figure~\ref{fig:res_donut} shows, for each fine-tuned version of Donut \cite{kim2022ocr}, the distribution of F1 scores over the RES. Since the embeddings were extracted from the pre-trained model and the fine-tuning process was carried out with the visual encoder frozen, the embedding map remains consistent across all versions.

\begin{figure}
\centering
\includegraphics[width=1\columnwidth]{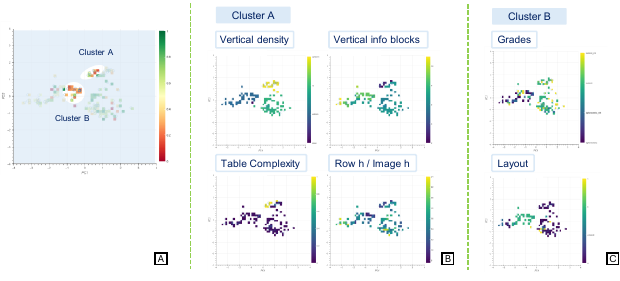}
\caption{Analysis of Donut’s \cite{kim2022ocr} Reduced Embedding Space (RES). Two low-performing regions are identified: \textit{region A}, characterized by a small number of information blocks and high table complexity, primarily affected by an external factor (the low zoom level); and \textit{region B}, driven by intrinsic document properties such as table structure and alphanumeric grading, but also affected by low zoom level.}
\label{fig:res_donut_w_features}
\end{figure}

Figure~\ref{fig:res_donut} reveals the emergence of distinct clusters in Donut's \cite{kim2022ocr} Reduced Embedding Space (RES) \footnote{VERSE is also applied to Idefics2 \cite{laurenccon2024matters} in \ref{verse_idefics2}.} This structure becomes evident once the model is fine-tuned on digital samples (Figure~\ref{fig:res_donut}.C). Some clusters consistently achieve higher F1 scores, while others perform systematically worse than the rest of the regions. Figure~\ref{fig:res_donut_w_features}.A highlights the two problematic regions identified for Donut \cite{kim2022ocr} (referred to as \textit{cluster A} and \textit{cluster B}).

In Figure~\ref{fig:res_donut_w_features}, \textit{cluster A} is characterized by a vertical distribution of information blocks located both at the top and bottom of the document, with a reduced number of blocks (two in total). However, although these characteristics help to understand which visual features Donut \cite{kim2022ocr} can distinguish in its RES, they are not the features inducing poor performance, as they are already present in the training dataset (starting from the digital version).

\textit{Cluster A} can also be discriminated based on the \textit{row\_h/image\_h} index (see definition in \ref{tab:metadata-condensed}), as this region corresponds to very low values of that index. We conduct the additional research in Figure \ref{fig:donut_zoom_research} to analyze how zoom affects model performance. From this analysis, it is concluded that the performance degradation in \textit{cluster A} is associated with an external feature of the document structure (namely, the zoom level).

Conversely, in addition to being located in a low-zoom region, \textit{cluster B} is segmented by intrinsic document characteristics, such as table structure and the way grades are represented. Specifically, the samples within this region exhibit alphanumeric grading systems (combining letters and numbers) and a template layout with two tables per page.

\subsubsection{Extrinsic document feature analysis: zoom level}
\label{subsec:extrinsic}

\begin{figure}
\centering
\includegraphics[width=0.8\columnwidth]{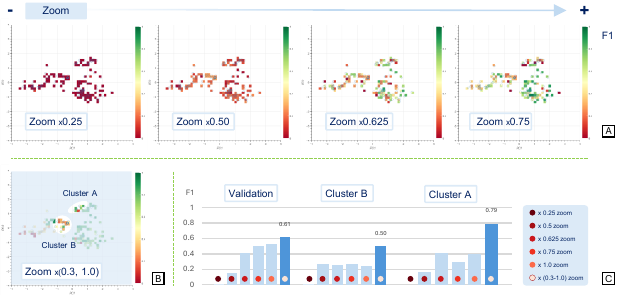}
\caption{Effect of training zoom levels on Donut’s \cite{kim2022ocr} performance. Fixed zoom-level experiments reveal that excessively low zoom values prevent the model from extracting textual information, while excessively high zoom values make the training task insufficiently challenging compared to the validation data. An intermediate zoom level (0.625) yields optimal results in the challenging areas \textit{cluster A} and \textit{cluster B} (A). Training with a uniform distribution of zoom values between 0.3 and 1.0 introduces visual diversity that enhances generalization (B). Comparative results across different regions (validation), \textit{cluster A}, and \textit{cluster B} (confirm that mixed zoom levels during fine-tuning consistently improve performance over fixed-zoom configurations (C)).}
\label{fig:donut_zoom_research}
\end{figure}

We evaluate Donut’s sensitivity to zoom during training using four different zoom levels: 0.25, 0.50, 0.625, and 0.75. Figure~\ref{fig:donut_zoom_research}.A illustrates the effect of the zoom level applied to the training samples on Donut’s inference performance. At low zoom levels, the model fails to learn effectively, as it cannot extract textual information from low-resolution images. As the zoom level increases, additional regions emerge in the Reduced Embedding Space (RES) that benefit from improved visual clarity. Focusing on \textit{cluster A}, the best performance is achieved at a zoom level of 0.625, while both lower (0.50) and higher (0.75) levels lead to performance degradation. This pattern suggests the existence of an optimal zoom range (sufficiently challenging to promote robust feature learning, yet not so extreme that the model fails to extract meaningful information).

Figure~\ref{fig:donut_zoom_research}.B presents an alternative approach. Instead of training with a fixed zoom level, we train with a uniform distribution of zoom values between 0.3 and 1.0. This variability introduces greater visual diversity in the training data, which proves beneficial for generalization. This strategy is particularly benefitial for \textit{clusters A} and \textit{B}.

We quantitatively compare the performance of these training strategies across three regions: the entire validation set, \textit{cluster A}, and \textit{cluster B}. Figure~\ref{fig:donut_zoom_research}.C shows that Donut \cite{kim2022ocr} achieves its best results across all regions when trained on samples with a distributed range of zoom levels (0.3–1.0). This behavior suggests that a significant portion of Donut’s \cite{kim2022ocr} strong performance on rendered samples (Figure~\ref{fig:res_donut}.F) can be attributed to the camera movements introduced during rendering, which naturally generate images with varying zoom levels.

\subsubsection{Intrinsic document features analysis}
\label{subsec:intrinsic}
We analyze the intrinsic characteristics of the documents, focusing on \textit{region B} of the Reduced Embedding Space (RES). This region was previously identified as problematic due to its prevalence of documents with dual tables and alphanumeric grading systems (intrinsic features), as well as low zoom levels (an extrinsic feature).

To address this limitation, Donut \cite{kim2022ocr} is fine-tuned using a specific subset of the MERIT Dataset (referred to as the \textit{booster set}) composed exclusively of samples exhibiting the two intrinsic features detected (this new MERIT split is not included in MERIT’s default training partition). This booster is used both independently and in combination with different base training sets.

\begin{figure}
\centering
\includegraphics[width=0.8\columnwidth]{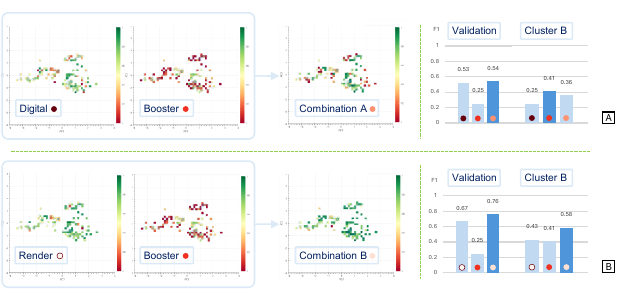}
\caption{Effect of training data composition on Donut’s \cite{kim2022ocr} performance. Fine-tuning with the \textit{booster set} (a subset containing intrinsic features) improves performance in the low-performing \textit{region B} (A). However, combining the booster with the digital dataset induces partial forgetting in that region while maintaining comparable generalization on the validation set. When the booster is combined with the rendered dataset (B), both \textit{region B} and the overall validation performance improve significantly. This complementary effect arises from jointly leveraging the intrinsic features of the booster and the zoom variability introduced by rendering.}
\label{fig:donut_intrinsic_research}
\end{figure}

Figure \ref{fig:donut_intrinsic_research}.A compares the results of fine-tuning Donut \cite{kim2022ocr} on the digital version of MERIT (Figure~\ref{fig:training_dataset}.A) versus the \textit{booster set}. The results show that the booster improves performance precisely in \textit{region B} (confirming that we have identified the key features governing this region). However, combining both datasets introduces a certain degree of forgetting in \textit{region B}, as the inclusion of less relevant features for this region dilutes its specificity (F1 decreases). Despite this, the combined model achieves a level of generalization on the validation set comparable to that of the digital model, although with slightly accentuated suboptimal behavior in regions previously under control.

In Figure \ref{fig:donut_intrinsic_research}.B, the same \textit{booster set} is combined with the rendered training set. In this case, the combination yields the most significant improvements (both in \textit{region B} and across the entire validation dataset). This behavior suggests a complementary effect: while the \textit{booster set} contributes the dual-table and alphanumeric-grade features, the rendered dataset introduces zoom variability (as previously discussed in Section \ref{subsec:extrinsic}), which enhances generalization. As a result, the model achieves a more balanced learning process by jointly incorporating the three characteristics previously identified as sources of low performance. This combination produces the best-performing Donut \cite{kim2022ocr} model across the entire validation set.

\section{Results}
\label{sec:results}
The results of VERSE can be analyzed along three main dimensions: (i) the intrinsic feasibility of each model for the target task, (ii) the performance gains achievable through clustering-guided fine-tuning, and (iii) the explainability and interpretability of dataset–model interactions.

\subsection{Model feasibility}
An analysis of the visual embedding space suggests that off-the-shelf models with limited clustering capabilities (reflected by lower Silhouette score values) tend to exhibit reduced performance in sequence generation tasks. Table~\ref{tab:results_triplet_loss_vs_f1} shows that PaliGemma \cite{beyer2024paligemma} and LLaVA \cite{liu2023visual} produce less structured embeddings and achieve lower F1 scores after fine-tuning. In contrast, Donut \cite{kim2022ocr} and Idefics2 \cite{laurenccon2024matters} yield more compact and discriminative embeddings, as indicated by higher Silhouette score values, and subsequently reach higher F1 scores.

\begin{table}[ht]
\centering
\scriptsize 
\begin{tabular}{lcccc}
\toprule
\makecell[tl]{\textbf{Model}\\\textbf{}} & 
\makecell[tl]{\textbf{Silhouette}\\\textbf{score}} & 
\makecell[tl]{\textbf{RES}\\\textbf{richness}} & 
\makecell[tl]{\textbf{}\\\textbf{}} & 
\makecell[tl]{\textbf{F1}\\\textbf{}} \\  
\midrule
Idefics2 \cite{laurenccon2024matters}& 0.63 & $\checkmark$ & $\Rightarrow$ &  0.8101\\
Donut \cite{kim2022ocr} & 0.50 & $\checkmark$ & $\Rightarrow$ & 0.7607\\
PaliGemma \cite{beyer2024paligemma} & 0.38 & $\times$ & $\Rightarrow$ & 0.3028\\
LLaVA \cite{liu2023visual} & 0.35 & $\times$ & $\Rightarrow$ & 0.0000\\
\bottomrule
\end{tabular}
\caption{Correlation between Silhouette score, RES richness, and downstream F1 for the best-performing model configurations.}
\label{tab:results_triplet_loss_vs_f1}
\end{table}

\subsection{Model Boosting}
We identify challenging clusters for suitable models (Donut \cite{kim2022ocr} and Idefics2 \cite{laurenccon2024matters}) and extract the corresponding features to target their underlying limitations. Both models exhibit difficulties in regions characterized by clusters containing alphanumeric scores, page layouts with two tables, and small visual attributes (e.g., zoomed-out content). Thus, we identify these features as the critical ones driving performance improvements for both models. Table~\ref{tab:model_boosting} reports the performance gains enabled by VERSE methodology.

\begin{table}[ht]
\centering
\tiny 
\begin{tabular}{llllllll}
\toprule
\makecell[tl]{\textbf{Model} } & 
\makecell[tl]{\textbf{Pre-VERSE}\\\textbf{F1}\tiny\textbf{(validation)}} & 
\makecell[tl]{\textbf{Region} } & 
\makecell[tl]{\textbf{F1}\\\tiny\textbf{(region)}} & 
\makecell[tl]{\textbf{Base}\\\textbf{dataset}} & 
\makecell[tl]{\textbf{Added boosting}\\\textbf{features}} & 
\makecell[tl]{$\Delta$\textbf{F1}\\\tiny\textbf{(validation)}} & 
\makecell[tl]{$\Delta$\textbf{F1}\\\tiny\textbf{(region)}} \\  
\midrule
\makecell[tl]{Donut\\\cite{kim2022ocr}} & 0.6712 & \textit{Cluster A} & 0.5989 & \textit{Render} & \makecell[tl]{Alphanumeric\\grading,\\double tables}  & \makecell[tl]{0.7607\\\tiny\textbf{(+0.09)}}  & \makecell[tl]{0.7828\\\tiny\textbf{(+0.18)}} \\
&  & \textit{Cluster B} & 0.4325 & &   &  & \makecell[tl]{0.5849\\\tiny\textbf{(+0.17)}} \\
\midrule
\makecell[tl]{Idefics2\\\cite{laurenccon2024matters}} & 0.7556 & \textit{Cluster A} & 0.5211 & \makecell[tl]{\textit{Digital}\\\textit{zoom}} 
 & Double tables & \makecell[tl]{0.8101\\\tiny\textbf{(+0.06)}}  & \makecell[tl]{0.6914\\\tiny\textbf{(+0.17)}}\\
&  & \textit{Cluster B} & 0.5054 &  &  &  & \makecell[tl]{0.6567\\\tiny\textbf{(+0.15)}}\\
&  & \textit{Cluster C} & 0.5040 &  &  &  & \makecell[tl]{0.7432\\\tiny\textbf{(+0.24)}}\\
\bottomrule
\end{tabular}
\caption{Performance improvements obtained after applying the VERSE methodology.
The \textit{Pre-VERSE F1} column reports baseline results obtained on the MERIT dataset, using human-centered selection criteria (\textit{render} images are the most photorealistic and thus have higher potential for model performance). 
The \textit{Region} column identifies model-specific clusters associated with low performance detected in the Reduced Embedding Space (RES).
Each experiment combines a base training dataset with an additional subset containing targeted \textit{boosting features}, designed to address weaknesses detected in those regions.
The resulting F1 improvements ($\Delta$F1) are shown for both the overall validation set and the specific target regions.
Results demonstrate that incorporating the identified boosting features leads to substantial performance gains in problematic clusters while preserving global generalization.}
\label{tab:model_boosting}
\end{table}

VERSE enhances performance on conflictive clusters without degrading performance on already well-generalized regions (thereby improving overall generalization). As a result, F1 scores increase across the entire MERIT Secret validation dataset. These improvements position the fine-tuned models at a level comparable to API-based solutions such as GPT4-O or Pixtral. Moreover, this gain comes with the additional benefits of working with local models, namely data privacy and governance, without compromising performance. Table~\ref{tab:results_comparison} summarizes the results of the best fine-tuned configurations for each model.

\begin{table}[ht]
\centering
\tiny
\begin{tabular}{lllc}
\toprule
\makecell[tl]{\textbf{Model} } & 
\makecell[tl]{\textbf{Deployment}} & 
\makecell[tl]{\textbf{Fine-tuned}} & 
\makecell[tl]{\textbf{F1}}\\
\midrule
Idefics2 \cite{laurenccon2024matters} & On-premise & VERSE & \textbf{0.8101}\\
GPT4-O & API-Based & API fine-tune & 0.7821\\
Donut \cite{kim2022ocr} & On-premise & VERSE & \textbf{0.7607} \\
Pixtral \cite{agrawal2024pixtral} & API-Based & N/A & 0.7267\\
\bottomrule
\end{tabular}
\caption{Comparison of the best-performing models. After applying the VERSE methodology, on-premise models achieve performance comparable to API-based solutions. Notably, Idefics2 \cite{laurenccon2024matters} fine-tuned with VERSE outperforms GPT4-O, demonstrating that locally deployed models can match (or even exceed) the accuracy of proprietary cloud services when efficiently optimized.}
\label{tab:results_comparison}
\end{table}

\subsection{Explainability}

The dimensionality reduction of the embedding space, combined with its graphical representation and the overlay of visual features, allows us to identify which document properties exert the greatest influence on the RES structure. In both Donut and Idefics2 (Figures \ref{fig:metadata_pca_donut} and \ref{fig:metadata_pca_idefics2}), we consistently observe that the features with the strongest impact on cluster formation and positioning are those intrinsic to the documents (namely, those that describe their macro-visual structure). These include the layout type, the vertical distribution of information, the number of table columns, the vertical position of the table on the page, or the number of vertical content blocks. All of these features directly shape the global organization of the document and, consequently, its representation in the RES.

In contrast, we argue that extrinsic features (those not inherent to the document’s structure) do not primarily determine the location of samples within the embedding space; rather, their distribution emerges as a consequence of the intrinsic characteristics that dominate each region. This is the case, for example, for zoom-related indices: although they do not govern the main spatial arrangement of samples in the RES, they remain relevant for explaining variations in performance (F1).

Similarly, the models do not appear to attend to visual features that are irrelevant to the table structure, such as the presence of signatures, stamps, or emblems (intrinsic but non-structural), nor to extrinsic artifacts, including shadows, creases, or other forms of noise in the document. Due to their limited relevance for the task and their weaker visual consistency, these features do not exert any observable influence on the organization of the RES.

\section{Discusion}
\label{sec:discusion}
In this section, we analyze the methodological implications and insights derived from applying VERSE to VrDU models, as well as its limitations and practical consequences.

Reducing the visual embedding space enables us to consistently analyze how different data-augmentation strategies affect the training samples within the RES. We observe that such techniques scatter the samples away from the clusters defined by their intrinsic and structural features, expanding their coverage over the manifold where real samples lie. This dispersion correlates with substantial improvements in F1 for both Donut \cite{kim2022ocr} and Idefics2 \cite{laurenccon2024matters}.

Another observation is that our results reinforce the idea that visual representation quality is critical for VrDU tasks. When the visual module fails to adequately disambiguate the document’s structure, the information forwarded to the textual decoder is already degraded, ultimately harming downstream performance. This perspective slightly diverges from the emphasis suggested by the authors of Idefics2 \cite{laurenccon2024matters}, who point to the textual decoder as the component with the greatest room for improvement. In contrast, our findings are more consistent with \cite{balestriero2025lejepa}, underscoring the importance of strong and task-aligned visual representations for solving multimodal tasks.

While VERSE provides valuable insight into explainability, we also note a significant degree of polysemanticity in the principal components. Although each PC cannot be interpreted as a clean combination of features, the qualitative analysis does reveal dominant patterns. The most problematic characteristics consistently correspond to low zoom levels, double-table layouts, and alphanumeric grading systems, which define the clusters with the lowest performance.

Finally, comparing models that share the same visual encoder (such as Idefics2 \cite{laurenccon2024matters} and PaliGemma \cite{beyer2024paligemma}, both based on SigLIP-So400M \cite{zhai2023sigmoid}) reveals that the encoder alone does not determine the richness of the visual embedding space. Idefics2 \cite{laurenccon2024matters} exhibits far greater sensitivity to VrDU-relevant features because its pre-training is more closely aligned with document-centric tasks. Additionally, its visual module is structurally more comprehensive, as it includes an internal MLP and Perceiver projection. As shown in Table~\ref{tab:embeddings_summary}, this architectural difference enables Idefics2 \cite{laurenccon2024matters} to produce richer and more task-aligned visual embeddings.

\section{Conclusions and contributions}
\label{sec:conclusions_contributions}

In this work, we presented VERSE, a methodology for reducing and exploring the visual embedding space of VLMs. By projecting the latent space and overlaying the relevant features that define the target samples, VERSE demonstrates how models internally structure visual information.

We further show that photorealistic synthetic data are not strictly required to improve performance. For training purposes, what matters is not human-perceived realism but whether synthetic samples populate regions of the visual–semantic embedding space that are meaningful and useful for the model.

Based on this methodology, our main contributions are:

\begin{itemize}
    \item Assessment of model feasibility for VrDU tasks, identifying which VLMs exhibit a sufficiently structured embedding space to handle the task effectively.

    \item Identification of error-inducing regions and determination of the visual features that must be reinforced in training data to mitigate them.

    \item Demonstration of targeted performance improvements: applying VERSE enhances low-performing regions without compromising overall generalization.
\end{itemize}

As future work, we plan to invert the VERSE pipeline: once problematic regions in the \textit{reduced embedding space} are identified, we aim to sample these areas directly and reconstruct images from their latent representations using generative models. This would enable the synthesis of data inherently aligned with the model’s latent space, providing a mechanism to generate training samples that embody the visual characteristics most relevant for improving model performance.

\section*{Data availability}
\label{sec:dataAvailability}

\begin{itemize}
    \item \textbf{Training:} The MERIT Dataset \cite{de2025merit} is available on \href{https://huggingface.co/datasets/de-Rodrigo/merit}{Hugging Face}. \footnote{MERIT Dataset on \href{https://huggingface.co/datasets/de-Rodrigo/merit}{Hugging Face}: https://huggingface.co/datasets/de-Rodrigo/merit}.
    \item \textbf{Validation:} The MERIT Secret Dataset is not available since it contains real samples under a Non-Disclosure Agreement.
    \item \textbf{Code:} The repository to train and test the models, and extract the embeddings from them is available on \href{https://github.com/nachoDRT/VrDU-Doctor/tree/main}{GitHub}. \footnote{Code on \href{https://github.com/nachoDRT/VrDU-Doctor/tree/main}{GitHub}: https://github.com/nachoDRT/VrDU-Doctor/tree/main}.
    \item \textbf{Training Sessions:} WandB links to training projects:
    \begin{itemize}
    \item \href{https://wandb.ai/ciclab-comillas/Donut}{Donut} \cite{kim2022ocr} \footnote{\href{https://wandb.ai/ciclab-comillas/Donut}{Donut} \cite{kim2022ocr} training project: https://wandb.ai/ciclab-comillas/Donut}.

    \item \href{https://wandb.ai/ciclab-comillas/Idefics2-patient}{Idefics2} \cite{laurenccon2024matters} \footnote{\href{https://wandb.ai/ciclab-comillas/Idefics2-patient}{Idefics2} \cite{laurenccon2024matters} training project : https://wandb.ai/ciclab-comillas/Idefics2-patient}.

    \item \href{https://wandb.ai/ciclab-comillas/Paligemma}{PaliGemma} \cite{beyer2024paligemma}\footnote{\href{https://wandb.ai/ciclab-comillas/Paligemma}{PaliGemma} \cite{beyer2024paligemma} training project: https://wandb.ai/ciclab-comillas/Paligemma}.

    \item \href{https://wandb.ai/ciclab-comillas/LLaVA}{LLaVA} \cite{liu2023visual} \footnote{\href{https://wandb.ai/ciclab-comillas/LLaVA}{LLaVA} \cite{liu2023visual} training project: https://wandb.ai/ciclab-comillas/LLaVA}.
\end{itemize}
\end{itemize}

\section*{Acknowledgments}
\label{sec:acknowledgments}
The authors of this publication would like to thank the Chair for Smart Industry for providing the necessary resources to produce this research. In addition, the authors extend their gratitude to the Secretary's Office staff at Universidad Pontificia Comillas for granting access to authentic transcripts of records under the condition of a Non-Disclosure Agreement, enabling the curation of MERIT Secret.

\appendix

\section{}
\label{MERIT_Secret_metadata}

Table \ref{tab:metadata-condensed} describes the features extracted from the validation dataset (MERIT Secret).

\begin{table}[ht]
\centering
\scriptsize 
\begin{tabular}{l p{9.5cm}}
\toprule
\textbf{Metadata} & \textbf{Description}\\  
\midrule
image\_w & Image width (px). \\
image\_h & Image height (px). \\
row\_w & Cell width (px). \\
row\_h & Cell height (px). \\
row\_h/image\_h & Cell-to-image height ratio. \\
row\_w/image\_w & Cell-to-image width ratio. \\
table\_complexity & Includes merged and split cells in a column. \\
grades\_system & Grading scale: Spanish (0–10), English (A*, A–U). \\
lacking\_grades & Missing grade(s) for a subject. \\
retake & Multiple grades for the same subject. \\
table\_grid & With/without grid lines (some only vertical). \\
grades & Numeric/alphanumeric, may include honors (MH, M). \\
orthogonal\_distortion & Misalignment or curved grids; excludes rotated but orthogonal tables. \\
white\_border & White border from scanning apps. \\
anonymization\_marks & Marks outside the table to ensure anonymity. \\
wrinkles & Wrinkled document. \\
v\_info\_blocks & Number of visual blocks (e.g., title, table, signature). \\
v\_density & Vertical distribution of blocks (uniform, top, bottom). \\
layout & Table type: A, A-double, B, C. \\
columns & Number of columns. \\
source & Image acquisition: camera or scanner. \\
shadows & Presence of shadows. \\
header\_badge & School emblem at top. \\
signed & Signed (signature may be anonymized). \\
stamped & School stamp. \\
table\_pos & Horizontal position: left, center, right. \\
\bottomrule
\end{tabular}
\caption{Features extracted from samples in MERIT Secret.}
\label{tab:metadata-condensed}
\end{table}

\section{}
\label{verse_idefics2}

\begin{figure}
\centering
\includegraphics[width=0.8\columnwidth]{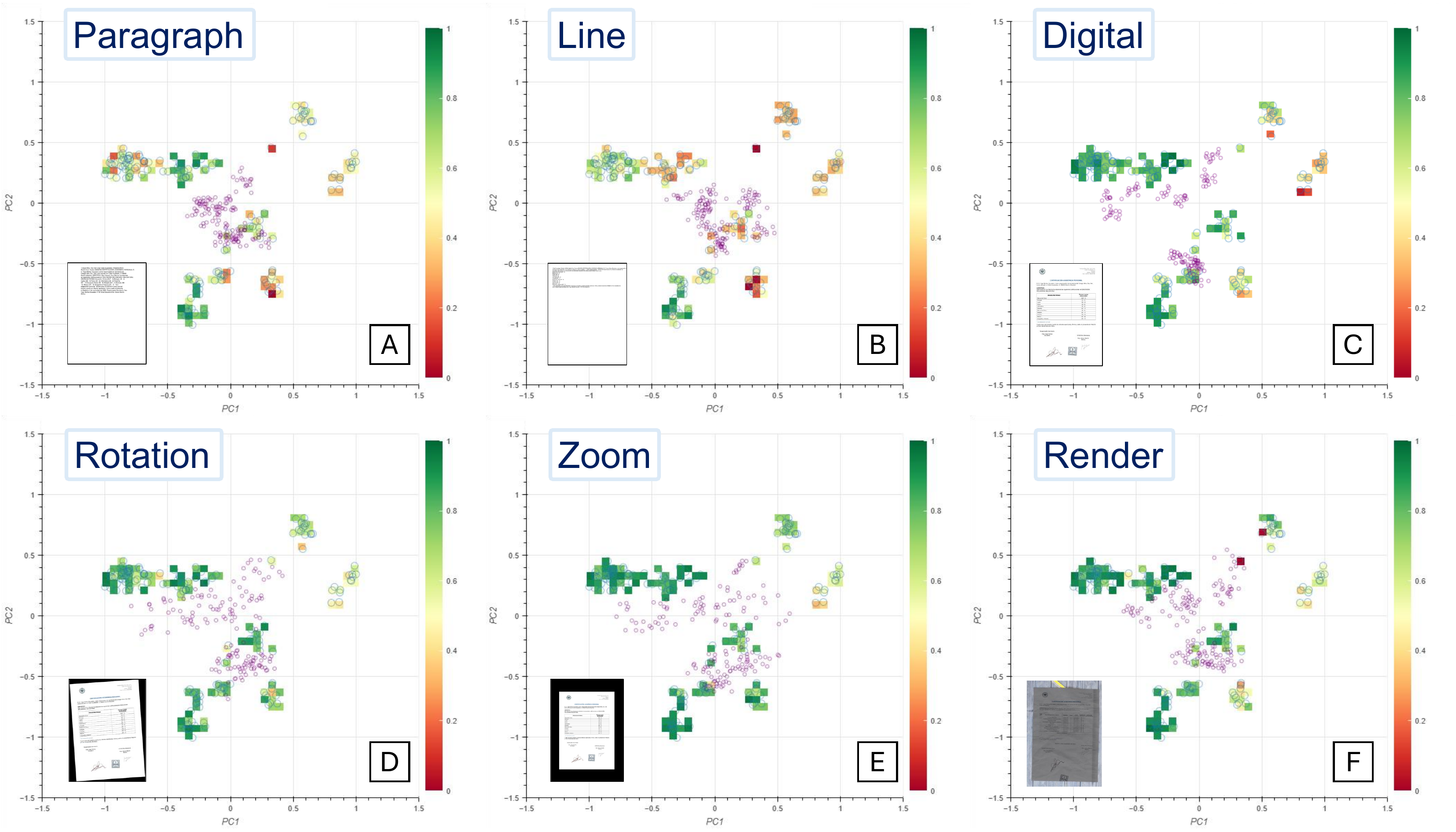}
\caption{Reduced Embedding Space (RES) for Idefics2 \cite{laurenccon2024matters} (PC1 vs. PC2). Higher F1 values are shown in green, while lower ones appear in red. Each subplot includes, in the lower-left corner, a representative training sample used for fine-tuning and a subset of training examples shown in purple. The visual richness of the training data increases progressively from A to F.}
\label{fig:res_idefics2}
\end{figure}

The application of VERSE to the Idefics2 \cite{laurenccon2024matters} model reveals that it does not only exhibits a more structured embedding space than Donut, but it can also obtain better results even when trained with limited levels of visual information. For instance, Idefics2 achieves strong performance in clusters composed of single-table documents, despite being trained on highly degraded versions of the trainin dataset (Figure~\ref{fig:res_idefics2}.A and B).

As additional visual features are introduced in the \textit{digital} version (\ref{fig:res_idefics2}.C), the training samples (shown in violet) shift toward more clearly defined clusters, indicating a good organization of the latent space. However, when data augmentation techniques are applied, the samples become more dispersed across the embedding space (a behavior that, although less pronounced, also appears in the rendered version, and correlates with better F1 scores).

\begin{figure}
\centering
\includegraphics[width=0.8\columnwidth]{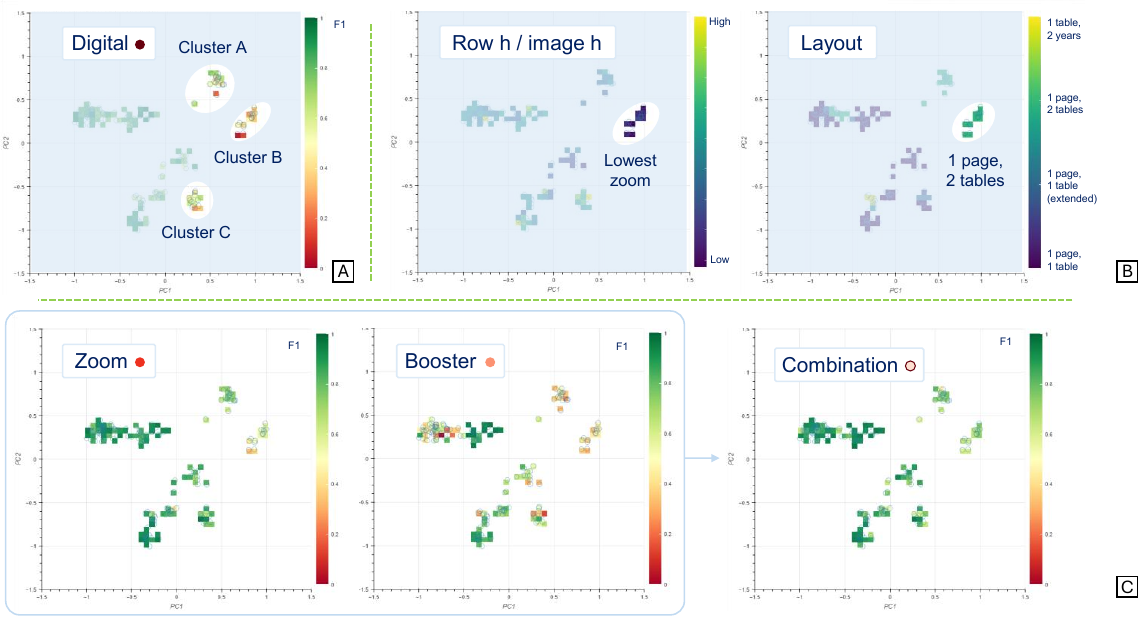}
\caption{Effect of training data composition on Idefics2’s~\cite{laurenccon2024matters} performance. \textit{Regions A}, \textit{B}, and \textit{C} are detected in the RES as conflictive clusters. Since \textit{Regions A} and \textit{C} already achieve solid performance when the model is fine-tuned on the \textit{zoom} version (C), we define \textit{region B} as the target cluster for the \textit{booster set}. VERSE suggests including samples with low zoom level and two tables as driving features of the \textit{booster set} (B). Fine-tuning the modle with the suggested combination improves performance in the problematic \textit{region B} while preserving consistent generalization across the \textit{validation} set and the remaining conflictive regions (C).}
\label{fig:idefics2_boosting}
\end{figure}

It is worth noting that Idefics2 performs particularly well in clusters dominated by single-table layouts. However, three low-performing regions persist, driven mainly by low zoom levels, dual-table layouts, and alphanumeric grading systems (Figure \ref{fig:idefics2_boosting}.A). Following the VERSE methodology, the most problematic area is identified as \textit{cluster B}, characterized by both a low zoom level and a double-table structure (Figure \ref{fig:idefics2_boosting}.B). When these features are included into the training set, the combination yields significant improvements over the entire validation set and achieves solid results in the previously problematic regions (Figure \ref{fig:idefics2_boosting}.C).

\begin{figure}
\centering
\includegraphics[width=0.8\columnwidth]{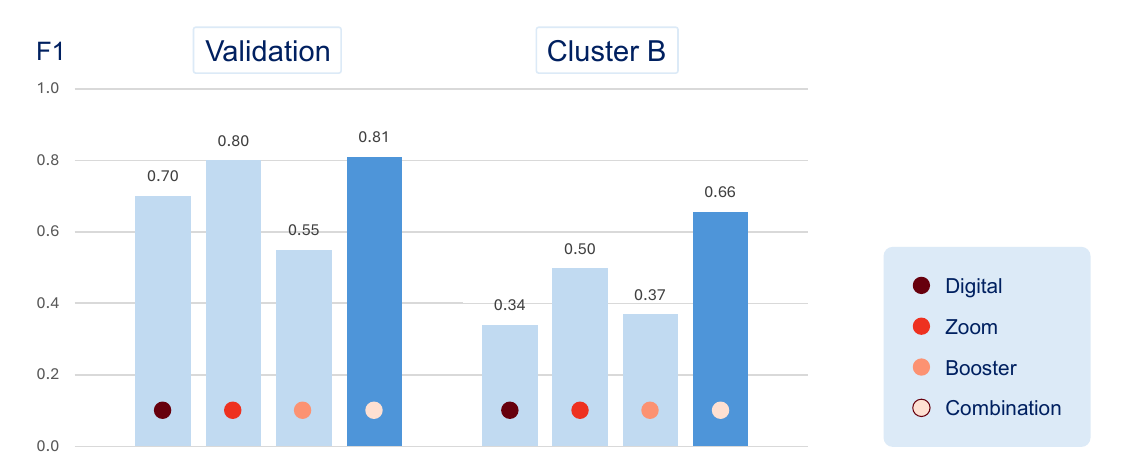}
\caption{Fine-tuning with the \textit{booster set} improves Idefics2’s~\cite{laurenccon2024matters} performance in the problematic \textit{region B} while preserving consistent generalization across the \textit{validation} set.}
\label{fig:idefics2_boosting_2}
\end{figure}

When we analyze the independent effects of each training set on the validation set and \textit{region B} (Figure~\ref{fig:idefics2_boosting_2}), the \textit{booster} dataset achieves slightly lower results in \textit{region B} than the model fine-tuned exclusively on the \textit{zoom} version. However, this behavior should be interpreted as a complementary learning effect rather than an overlap: each dataset contributes distinct strengths (the zoom version enhances the model’s ability to extract meaningful information from low-resolution tables, while the booster improves its understanding of double-table layouts). When combined, these two effects reinforce each other, resulting in better overall performance than either could achieve independently.

\bibliographystyle{elsarticle-num-names} 
\bibliography{vrdu-doctor-bibliography}

@Article{de2025merit,
  author    = {De Rodrigo, Ignacio and Sanchez-Cuadrado, Alberto and Boal, Jaime and Lopez-Lopez, Alvaro J},
  journal   = {Pattern Recognition},
  title     = {The MERIT dataset: Modelling and efficiently rendering interpretable transcripts},
  year      = {2025},
  pages     = {112502},
  publisher = {Elsevier},
}

@Article{laurenccon2024matters,
  author  = {Lauren{\c{c}}on, Hugo and Tronchon, L{\'e}o and Cord, Matthieu and Sanh, Victor},
  journal = {Advances in Neural Information Processing Systems},
  title   = {What matters when building vision-language models?},
  year    = {2024},
  pages   = {87874--87907},
  volume  = {37},
}

@InProceedings{kim2022ocr,
  author       = {Kim, Geewook and Hong, Teakgyu and Yim, Moonbin and Nam, JeongYeon and Park, Jinyoung and Yim, Jinyeong and Hwang, Wonseok and Yun, Sangdoo and Han, Dongyoon and Park, Seunghyun},
  booktitle    = {European Conference on Computer Vision},
  title        = {Ocr-free document understanding transformer},
  year         = {2022},
  organization = {Springer},
  pages        = {498--517},
}

@Article{beyer2024paligemma,
  author  = {Beyer, Lucas and Steiner, Andreas and Pinto, Andr{\'e} Susano and Kolesnikov, Alexander and Wang, Xiao and Salz, Daniel and Neumann, Maxim and Alabdulmohsin, Ibrahim and Tschannen, Michael and Bugliarello, Emanuele and others},
  journal = {arXiv preprint arXiv:2407.07726},
  title   = {Paligemma: A versatile 3b vlm for transfer},
  year    = {2024},
}

@Article{liu2023visual,
  author  = {Liu, Haotian and Li, Chunyuan and Wu, Qingyang and Lee, Yong Jae},
  journal = {Advances in neural information processing systems},
  title   = {Visual instruction tuning},
  year    = {2023},
  pages   = {34892--34916},
  volume  = {36},
}

@InProceedings{livathinos2025docling,
  author    = {Livathinos, Nikos and Auer, Christoph and Lysak, Maxim and Nassar, Ahmed and Dolfi, Michele and Vagenas, Panos and Ramis, Cesar Berrospi and Omenetti, Matteo and Dinkla, Kasper and Kim, Yusik and others},
  booktitle = {AAAI Conference on Artificial Intelligence},
  title     = {Docling: An Efficient Open-Source Toolkit for AI-driven Document Conversion},
  year      = {2025},
}

@InProceedings{xu2020layoutlm,
  author    = {Xu, Yiheng and Li, Minghao and Cui, Lei and Huang, Shaohan and Wei, Furu and Zhou, Ming},
  booktitle = {Proceedings of the 26th ACM SIGKDD international conference on knowledge discovery \& data mining},
  title     = {Layoutlm: Pre-training of text and layout for document image understanding},
  year      = {2020},
  pages     = {1192--1200},
}

@InProceedings{xu2021layoutlmv2,
  author    = {Xu, Yang and Xu, Yiheng and Lv, Tengchao and Cui, Lei and Wei, Furu and Wang, Guoxin and Lu, Yijuan and Florencio, Dinei and Zhang, Cha and Che, Wanxiang and others},
  booktitle = {Proceedings of the 59th Annual Meeting of the Association for Computational Linguistics and the 11th International Joint Conference on Natural Language Processing (Volume 1: Long Papers)},
  title     = {Layoutlmv2: Multi-modal pre-training for visually-rich document understanding},
  year      = {2021},
  pages     = {2579--2591},
}

@InProceedings{huang2022layoutlmv3,
  author    = {Huang, Yupan and Lv, Tengchao and Cui, Lei and Lu, Yutong and Wei, Furu},
  booktitle = {Proceedings of the 30th ACM International Conference on Multimedia},
  title     = {Layoutlmv3: Pre-training for document ai with unified text and image masking},
  year      = {2022},
  pages     = {4083--4091},
}

@Article{Xu2021,
  author  = {Xu, Yiheng and Lv, Tengchao and Cui, Lei and Wang, Guoxin and Lu, Yijuan and Florencio, Dinei and Zhang, Cha and Wei, Furu},
  journal = {arXiv preprint arXiv:2104.08836},
  title   = {Layoutxlm: Multimodal pre-training for multilingual visually-rich document understanding},
  year    = {2021},
}

@InProceedings{Xu2022,
  author    = {Xu, Yiheng and Lv, Tengchao and Cui, Lei and Wang, Guoxin and Lu, Yijuan and Florencio, Dinei and Zhang, Cha and Wei, Furu},
  booktitle = {Findings of the Association for Computational Linguistics: ACL 2022},
  title     = {XFUND: a benchmark dataset for multilingual visually rich form understanding},
  year      = {2022},
  pages     = {3214--3224},
}

@Article{palacios2008system,
  author    = {Palacios, Rafael and Gupta, Amar},
  journal   = {Image and Vision Computing},
  title     = {A system for processing handwritten bank checks automatically},
  year      = {2008},
  number    = {10},
  pages     = {1297--1313},
  volume    = {26},
  publisher = {Elsevier},
}

@InProceedings{gu2022xylayoutlm,
  author    = {Gu, Zhangxuan and Meng, Changhua and Wang, Ke and Lan, Jun and Wang, Weiqiang and Gu, Ming and Zhang, Liqing},
  booktitle = {Proceedings of the IEEE/CVF conference on computer vision and pattern recognition},
  title     = {Xylayoutlm: Towards layout-aware multimodal networks for visually-rich document understanding},
  year      = {2022},
  pages     = {4583--4592},
}

@InProceedings{dhouib2023docparser,
  author       = {Dhouib, Mohamed and Bettaieb, Ghassen and Shabou, Aymen},
  booktitle    = {International Conference on Document Analysis and Recognition},
  title        = {Docparser: End-to-end ocr-free information extraction from visually rich documents},
  year         = {2023},
  organization = {Springer},
  pages        = {155--172},
}

@InProceedings{tang2023unifying,
  author    = {Tang, Zineng and Yang, Ziyi and Wang, Guoxin and Fang, Yuwei and Liu, Yang and Zhu, Chenguang and Zeng, Michael and Zhang, Cha and Bansal, Mohit},
  booktitle = {Proceedings of the IEEE/CVF Conference on Computer Vision and Pattern Recognition},
  title     = {Unifying vision, text, and layout for universal document processing},
  year      = {2023},
  pages     = {19254--19264},
}

@InProceedings{Jaume2019,
  author       = {Jaume, Guillaume and Ekenel, Hazim Kemal and Thiran, Jean-Philippe},
  booktitle    = {2019 International Conference on Document Analysis and Recognition Workshops (ICDARW)},
  title        = {Funsd: A dataset for form understanding in noisy scanned documents},
  year         = {2019},
  organization = {IEEE},
  pages        = {1--6},
  volume       = {2},
}

@InProceedings{ringland2019nne,
  author       = {Ringland, Nicky and Dai, Xiang and Hachey, Ben and Karimi, Sarvnaz and Paris, Cecile and Curran, James R},
  booktitle    = {Proceedings of the 57th Annual Meeting of the Association for Computational Linguistics},
  title        = {NNE: A Dataset for Nested Named Entity Recognition in English Newswire},
  year         = {2019},
  organization = {Association for Computational Linguistics},
}

@InProceedings{Park2019,
  author    = {Park, Seunghyun and Shin, Seung and Lee, Bado and Lee, Junyeop and Surh, Jaeheung and Seo, Minjoon and Lee, Hwalsuk},
  booktitle = {Workshop on Document Intelligence at NeurIPS 2019},
  title     = {CORD: a consolidated receipt dataset for post-OCR parsing},
  year      = {2019},
}

@InProceedings{huang2019icdar2019,
  author       = {Huang, Zheng and Chen, Kai and He, Jianhua and Bai, Xiang and Karatzas, Dimosthenis and Lu, Shijian and Jawahar, CV},
  booktitle    = {2019 International Conference on Document Analysis and Recognition (ICDAR)},
  title        = {Icdar2019 competition on scanned receipt ocr and information extraction},
  year         = {2019},
  organization = {IEEE},
  pages        = {1516--1520},
}

@InProceedings{zhong2019publaynet,
  author       = {Zhong, Xu and Tang, Jianbin and Yepes, Antonio Jimeno},
  booktitle    = {2019 International conference on document analysis and recognition (ICDAR)},
  title        = {Publaynet: largest dataset ever for document layout analysis},
  year         = {2019},
  organization = {IEEE},
  pages        = {1015--1022},
}

@InProceedings{mathew2021docvqa,
  author    = {Mathew, Minesh and Karatzas, Dimosthenis and Jawahar, CV},
  booktitle = {Proceedings of the IEEE/CVF winter conference on applications of computer vision},
  title     = {Docvqa: A dataset for vqa on document images},
  year      = {2021},
  pages     = {2200--2209},
}

@InProceedings{tanaka2023slidevqa,
  author    = {Tanaka, Ryota and Nishida, Kyosuke and Nishida, Kosuke and Hasegawa, Taku and Saito, Itsumi and Saito, Kuniko},
  booktitle = {Proceedings of the AAAI Conference on Artificial Intelligence},
  title     = {Slidevqa: A dataset for document visual question answering on multiple images},
  year      = {2023},
  number    = {11},
  pages     = {13636--13645},
  volume    = {37},
}

@InProceedings{mathew2022infographicvqa,
  author    = {Mathew, Minesh and Bagal, Viraj and Tito, Rub{\`e}n and Karatzas, Dimosthenis and Valveny, Ernest and Jawahar, CV},
  booktitle = {Proceedings of the IEEE/CVF Winter Conference on Applications of Computer Vision},
  title     = {Infographicvqa},
  year      = {2022},
  pages     = {1697--1706},
}

@Article{agrawal2024pixtral,
  author  = {Agrawal, Pravesh and Antoniak, Szymon and Hanna, Emma Bou and Bout, Baptiste and Chaplot, Devendra and Chudnovsky, Jessica and Costa, Diogo and De Monicault, Baudouin and Garg, Saurabh and Gervet, Theophile and others},
  journal = {arXiv preprint arXiv:2410.07073},
  title   = {Pixtral 12B},
  year    = {2024},
}

@InProceedings{ding2023vqa,
  author       = {Ding, Yihao and Luo, Siwen and Chung, Hyunsuk and Han, Soyeon Caren},
  booktitle    = {Joint European Conference on Machine Learning and Knowledge Discovery in Databases},
  title        = {VQA: A New Dataset for Real-World VQA on PDF Documents},
  year         = {2023},
  organization = {Springer},
  pages        = {585--601},
}

@Article{laurenccon2024building,
  author  = {Lauren{\c{c}}on, Hugo and Marafioti, Andr{\'e}s and Sanh, Victor and Tronchon, L{\'e}o},
  journal = {arXiv preprint arXiv:2408.12637},
  title   = {Building and better understanding vision-language models: insights and future directions},
  year    = {2024},
}

@InProceedings{lysak2023optimized,
  author       = {Lysak, Maksym and Nassar, Ahmed and Livathinos, Nikolaos and Auer, Christoph and Staar, Peter},
  booktitle    = {International Conference on Document Analysis and Recognition},
  title        = {Optimized table tokenization for table structure recognition},
  year         = {2023},
  organization = {Springer},
  pages        = {37--50},
}

@Article{nassar2025smoldocling,
  author  = {Nassar, Ahmed and Marafioti, Andres and Omenetti, Matteo and Lysak, Maksym and Livathinos, Nikolaos and Auer, Christoph and Morin, Lucas and de Lima, Rafael Teixeira and Kim, Yusik and Gurbuz, A Said and others},
  journal = {arXiv preprint arXiv:2503.11576},
  title   = {SmolDocling: An ultra-compact vision-language model for end-to-end multi-modal document conversion},
  year    = {2025},
}

@InProceedings{ziegler2024craft,
  author    = {Ziegler, Ingo and K{\"o}ksal, Abdullatif and Elliott, Desmond and Schuetze, Hinrich},
  booktitle = {NeurIPS 2024 Workshop on Fine-Tuning in Modern Machine Learning: Principles and Scalability},
  title     = {CRAFT Your Dataset: Task-Specific Synthetic Dataset Generation Through Corpus Retrieval and Augmentation},
  year      = {2024},
}

@InProceedings{chenpali,
  author    = {Chen, Xi and Wang, Xiao and Changpinyo, Soravit and Piergiovanni, AJ and Padlewski, Piotr and Salz, Daniel and Goodman, Sebastian and Grycner, Adam and Mustafa, Basil and Beyer, Lucas and others},
  booktitle = {The Eleventh International Conference on Learning Representations},
  title     = {PaLI: A Jointly-Scaled Multilingual Language-Image Model},
  year      = {2023},
}

@InProceedings{zhai2023sigmoid,
  author    = {Zhai, Xiaohua and Mustafa, Basil and Kolesnikov, Alexander and Beyer, Lucas},
  booktitle = {Proceedings of the IEEE/CVF international conference on computer vision},
  title     = {Sigmoid loss for language image pre-training},
  year      = {2023},
  pages     = {11975--11986},
}

@Article{team2024gemma,
  author  = {Team, Gemma and Mesnard, Thomas and Hardin, Cassidy and Dadashi, Robert and Bhupatiraju, Surya and Pathak, Shreya and Sifre, Laurent and Rivi{\`e}re, Morgane and Kale, Mihir Sanjay and Love, Juliette and others},
  journal = {arXiv preprint arXiv:2403.08295},
  title   = {Gemma: Open models based on gemini research and technology},
  year    = {2024},
}

@InProceedings{radford2021learning,
  author       = {Radford, Alec and Kim, Jong Wook and Hallacy, Chris and Ramesh, Aditya and Goh, Gabriel and Agarwal, Sandhini and Sastry, Girish and Askell, Amanda and Mishkin, Pamela and Clark, Jack and others},
  booktitle    = {International conference on machine learning},
  title        = {Learning transferable visual models from natural language supervision},
  year         = {2021},
  organization = {PmLR},
  pages        = {8748--8763},
}

@Misc{vicuna2023,
  author = {Chiang, Wei-Lin and Li, Zhuohan and Lin, Zi and Sheng, Ying and Wu, Zhanghao and Zhang, Hao and Zheng, Lianmin and Zhuang, Siyuan and Zhuang, Yonghao and Gonzalez, Joseph E. and Stoica, Ion and Xing, Eric P.},
  month  = {March},
  title  = {Vicuna: An Open-Source Chatbot Impressing GPT-4 with 90\%* ChatGPT Quality},
  year   = {2023},
  url    = {https://lmsys.org/blog/2023-03-30-vicuna/},
}

@InProceedings{spies2025transformers,
  author    = {Spies, Alexander F and Edwards, William and Ivanitskiy, Michael and Skapars, Adrians and R{\"a}uker, Tilman and Inoue, Katsumi and Russo, Alessandra and Shanahan, Murray},
  booktitle = {ICLR 2025 Workshop on World Models: Understanding, Modelling and Scaling},
  title     = {Transformers Use Causal World Models in Maze-Solving Tasks},
  year      = {2024},
}

@Article{li2025multi,
  author    = {Li, Zheng and Guo, Caili and Wang, Xin and Zhang, Hao and Hu, Lin},
  journal   = {Pattern Recognition},
  title     = {Multi-view visual semantic embedding for cross-modal image--text retrieval},
  year      = {2025},
  pages     = {111088},
  volume    = {159},
  publisher = {Elsevier},
}

@InProceedings{liu2021swin,
  author    = {Liu, Ze and Lin, Yutong and Cao, Yue and Hu, Han and Wei, Yixuan and Zhang, Zheng and Lin, Stephen and Guo, Baining},
  booktitle = {Proceedings of the IEEE/CVF international conference on computer vision},
  title     = {Swin transformer: Hierarchical vision transformer using shifted windows},
  year      = {2021},
  pages     = {10012--10022},
}

@Article{balestriero2025lejepa,
  author  = {Balestriero, Randall and LeCun, Yann},
  journal = {arXiv preprint arXiv:2511.08544},
  title   = {LeJEPA: Provable and Scalable Self-Supervised Learning Without the Heuristics},
  year    = {2025},
}

@Article{laurenccon2023obelics,
  author  = {Lauren{\c{c}}on, Hugo and Saulnier, Lucile and Tronchon, L{\'e}o and Bekman, Stas and Singh, Amanpreet and Lozhkov, Anton and Wang, Thomas and Karamcheti, Siddharth and Rush, Alexander and Kiela, Douwe and others},
  journal = {Advances in Neural Information Processing Systems},
  title   = {Obelics: An open web-scale filtered dataset of interleaved image-text documents},
  year    = {2023},
  pages   = {71683--71702},
  volume  = {36},
}

@InProceedings{biten2022ocr,
  author       = {Biten, Ali Furkan and Tito, Ruben and Gomez, Lluis and Valveny, Ernest and Karatzas, Dimosthenis},
  booktitle    = {European Conference on Computer Vision},
  title        = {Ocr-idl: Ocr annotations for industry document library dataset},
  year         = {2022},
  organization = {Springer},
  pages        = {241--252},
}

@InProceedings{sharma2018conceptual,
  author    = {Sharma, Piyush and Ding, Nan and Goodman, Sebastian and Soricut, Radu},
  booktitle = {Proceedings of the 56th Annual Meeting of the Association for Computational Linguistics (Volume 1: Long Papers)},
  title     = {Conceptual captions: A cleaned, hypernymed, image alt-text dataset for automatic image captioning},
  year      = {2018},
  pages     = {2556--2565},
}

@InProceedings{changpinyo2022all,
  author    = {Changpinyo, Soravit and Kukliansy, Doron and Szpektor, Idan and Chen, Xi and Ding, Nan and Soricut, Radu},
  booktitle = {Proceedings of the 2022 conference of the north american chapter of the association for computational linguistics: human language technologies},
  title     = {All you may need for VQA are image captions},
  year      = {2022},
  pages     = {1947--1963},
}

@Article{kuznetsova2020open,
  author    = {Kuznetsova, Alina and Rom, Hassan and Alldrin, Neil and Uijlings, Jasper and Krasin, Ivan and Pont-Tuset, Jordi and Kamali, Shahab and Popov, Stefan and Malloci, Matteo and Kolesnikov, Alexander and others},
  journal   = {International journal of computer vision},
  title     = {The open images dataset v4: Unified image classification, object detection, and visual relationship detection at scale},
  year      = {2020},
  number    = {7},
  pages     = {1956--1981},
  volume    = {128},
  publisher = {Springer},
}

@InProceedings{srinivasan2021wit,
  author    = {Srinivasan, Krishna and Raman, Karthik and Chen, Jiecao and Bendersky, Michael and Najork, Marc},
  booktitle = {Proceedings of the 44th international ACM SIGIR conference on research and development in information retrieval},
  title     = {Wit: Wikipedia-based image text dataset for multimodal multilingual machine learning},
  year      = {2021},
  pages     = {2443--2449},
}





\end{document}